\documentclass[10pt,twocolumn,letterpaper]{article}

\usepackage{cvpr}
\usepackage{times}
\usepackage{epsfig}
\usepackage{graphicx}
\usepackage{amsmath}
\usepackage{amssymb}
\usepackage{balance}
\usepackage{url}
\usepackage{mathtools}
\usepackage{xspace}
\usepackage{multirow}
\usepackage{booktabs}
\usepackage{algorithm}
\usepackage{algorithmic}
\usepackage{xcolor}
\usepackage{cite}  

\graphicspath{{./}{./figures/}}

\def \R{{\mathbb{R}}}
\def \S{{\mathcal{S}}}
\def \X{{\mathcal{X}}}
\def \N{{\mathcal{N}}}

\def \D{{\mathcal{D}}}
\def \P{{\mathcal{P}}}

\def \phit{{\tilde{\phi}}}
\def \Gammat{{\tilde{\Gamma}}}

\def \IoU{{\textnormal{IoU}}}

\newcommand{\ra}[1]{\renewcommand{\arraystretch}{#1}}

\usepackage{amsthm}

\usepackage[pagebackref=true,breaklinks=true,letterpaper=true,colorlinks,bookmarks=false]{hyperref}

\cvprfinalcopy 

\def\cvprPaperID{} 

\ifcvprfinal\fi
\begin{document}

\title{Deep Level Sets: Implicit Surface Representations for 3D Shape Inference}

\author{
Mateusz Michalkiewicz$^1$, 
Jhony K. Pontes$^1$, 
Dominic Jack$^1$, 
Mahsa Baktashmotlagh$^2$,
Anders Eriksson$^2$ \vspace{3mm} \\
$^1$School of Electrical Engineering and Computer Science, Queensland University of Technology\\
$^2$School of Information Technology and Electrical Engineering, University of Queensland
}

\maketitle

\begin{abstract}
\end{abstract}
Existing 3D surface representation approaches are unable to accurately classify pixels and their orientation lying on the boundary of an object. Thus resulting in coarse representations which usually require post-processing steps to extract 3D surface meshes. To overcome this limitation, we propose an end-to-end trainable model that directly predicts implicit surface representations of arbitrary topology by optimising a novel geometric loss function. 
Specifically, we propose to represent the output as an oriented level set of a continuous embedding function, and incorporate this in a deep end-to-end learning framework by introducing a variational shape inference formulation. 
We investigate the benefits of our approach on the task of 3D surface prediction and demonstrate its ability to produce a more accurate reconstruction compared to voxel-based representations. We further show that our model is flexible and can be applied to a variety of shape inference problems.

\section{Introduction}
\label{sec:intro}
In recent years, the field of 3D reconstruction has achieved great progress trying to tackle many categories of problems such as structure from motion \cite{haming2010structure}, multi-view stereo \cite{furukawa2010towards} and reconstruction from a single image \cite{criminisi2000single}. The application domain includes, but is not limited to, robotic-assisted surgery, self-driving cars, intelligent robots, and helping visually impaired people to interact with the surrounding world via augmented reality.

\begin{figure}[t]
    \centering
    \includegraphics[width=.2\textwidth]{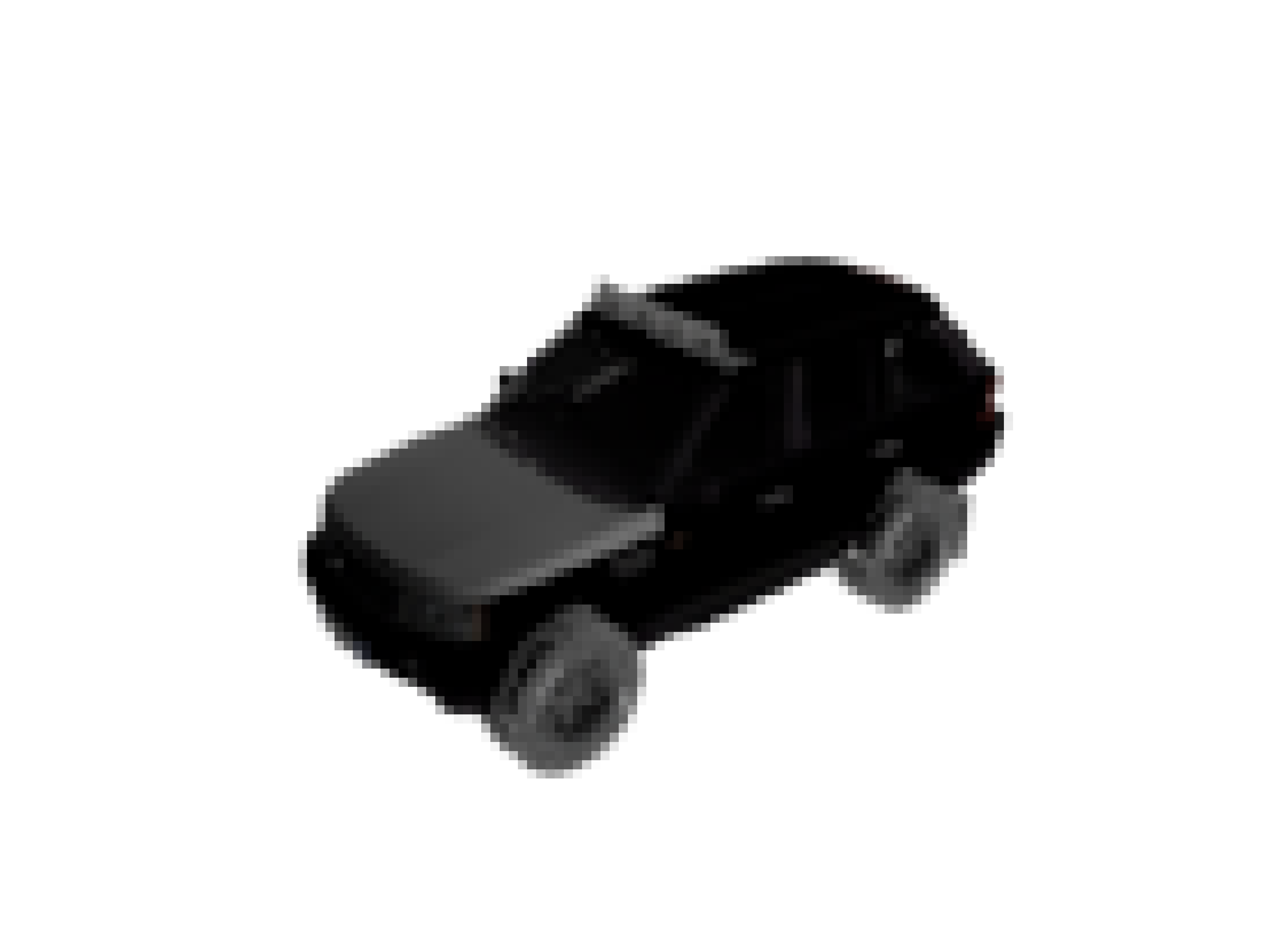}\hspace{3mm} 
    \includegraphics[width=.2\textwidth]{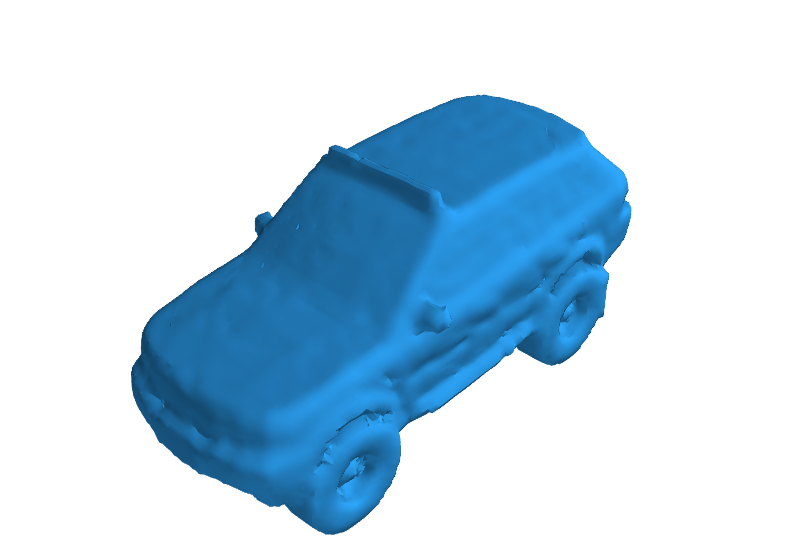}\hspace{3mm} \\
    \vspace{3mm}
    \includegraphics[width=.2\textwidth]{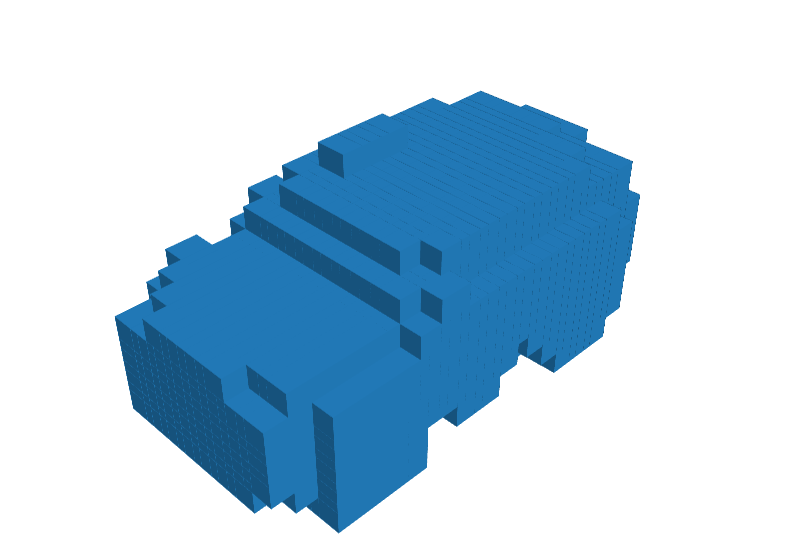}\hspace{3mm}
    \includegraphics[width=.2\textwidth]{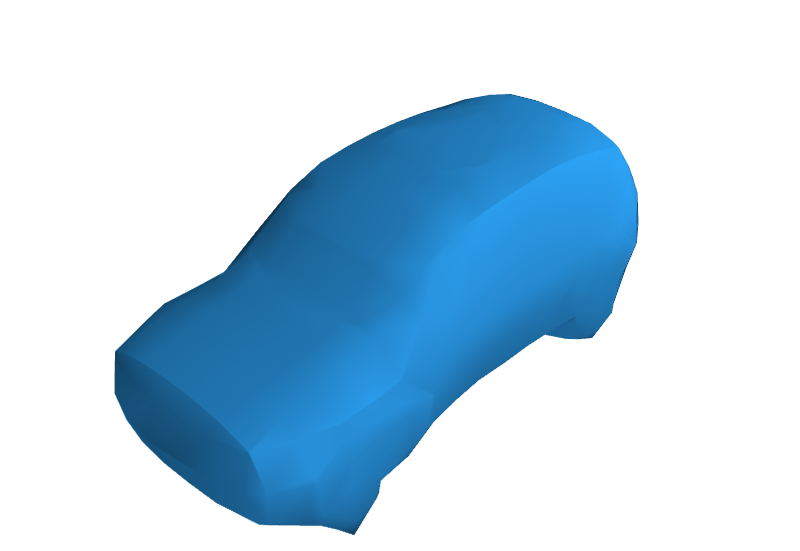}
    \caption{3D shape inference from a single image. 
    (top left) Input image, (top right) ground-truth, 
    (bottom left) predicted shape using a volumetric representation, 
    (bottom right) predicted shape using our proposed implicit shape 
    representation. Both representations have a resolution of $20^3$.
    } 
    \label{fig:teaser}
\end{figure}

A majority of existing 3D representation learning approaches are based 
on voxel occupancy 
\cite{girdhar2016learning,choy20163d,rezende2016unsupervised,richter2018matryoshka} 
 but a considerable amount of attention has also 
been put on point clouds \cite{qi2017pointnet,fan2017point} 
and explicit shape parameterisation \cite{liao2018deep}. 

\begin{figure*}[t]
    \centering
    \begin{minipage}[t]{.08\textwidth}
    \vspace{-14mm}
    {\footnotesize
	\textcolor{red}{- Irregular} \\
    \textcolor{green}{+ Geometry} \\
    \textcolor{red}{- Learning} }
    \end{minipage}
    \includegraphics[width=0.12\textwidth]{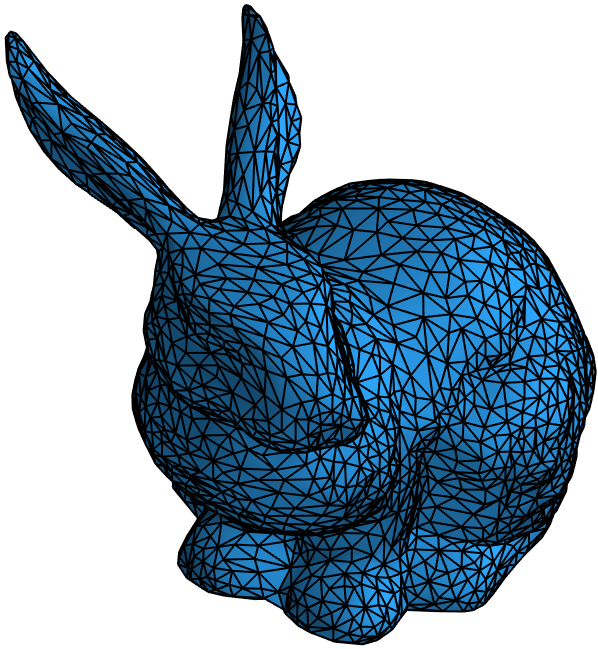} \qquad 
    \begin{minipage}[t]{.08\textwidth}
    \vspace{-14mm}
    {\footnotesize
	\textcolor{green}{+ Regular} \\
    \textcolor{red}{- Geometry} \\
    \textcolor{green}{+ Learning}}
    \end{minipage}
    \includegraphics[width=0.12\textwidth]{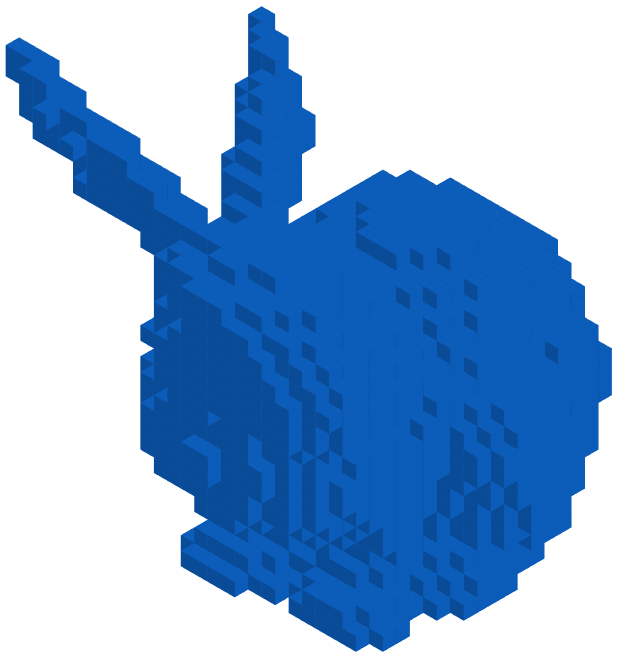} \qquad 
    \begin{minipage}[t]{.08\textwidth}
    \vspace{-14mm}
    {\footnotesize
	\textcolor{red}{- Irregular} \\
    \textcolor{orange}{$\pm$Geometry} \\
    \textcolor{red}{- Learning} 
    }
    \end{minipage}
     \includegraphics[width=0.12\textwidth]{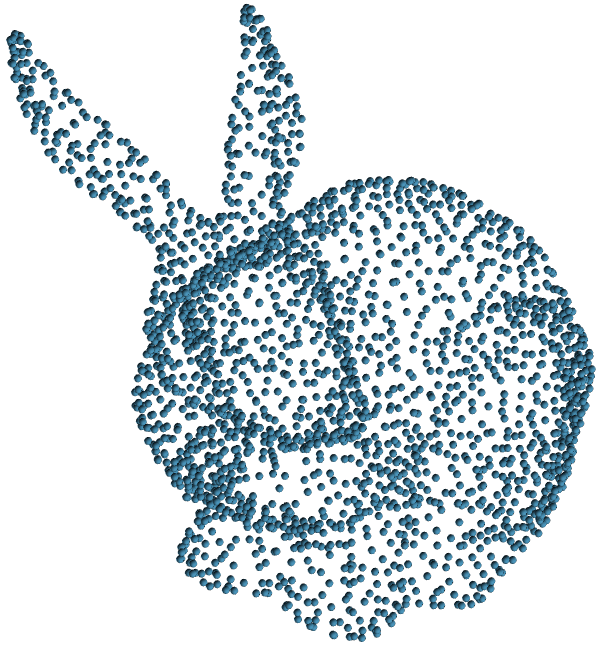} \qquad 
    \begin{minipage}[t]{.08\textwidth}
    \vspace{-14mm}
    {\footnotesize
	\textcolor{green}{+ Regular} \\
    \textcolor{green}{+ Geometry} \\
    \textcolor{green}{+ Learning}
    }
    \end{minipage}
    \includegraphics[width=0.15\textwidth]{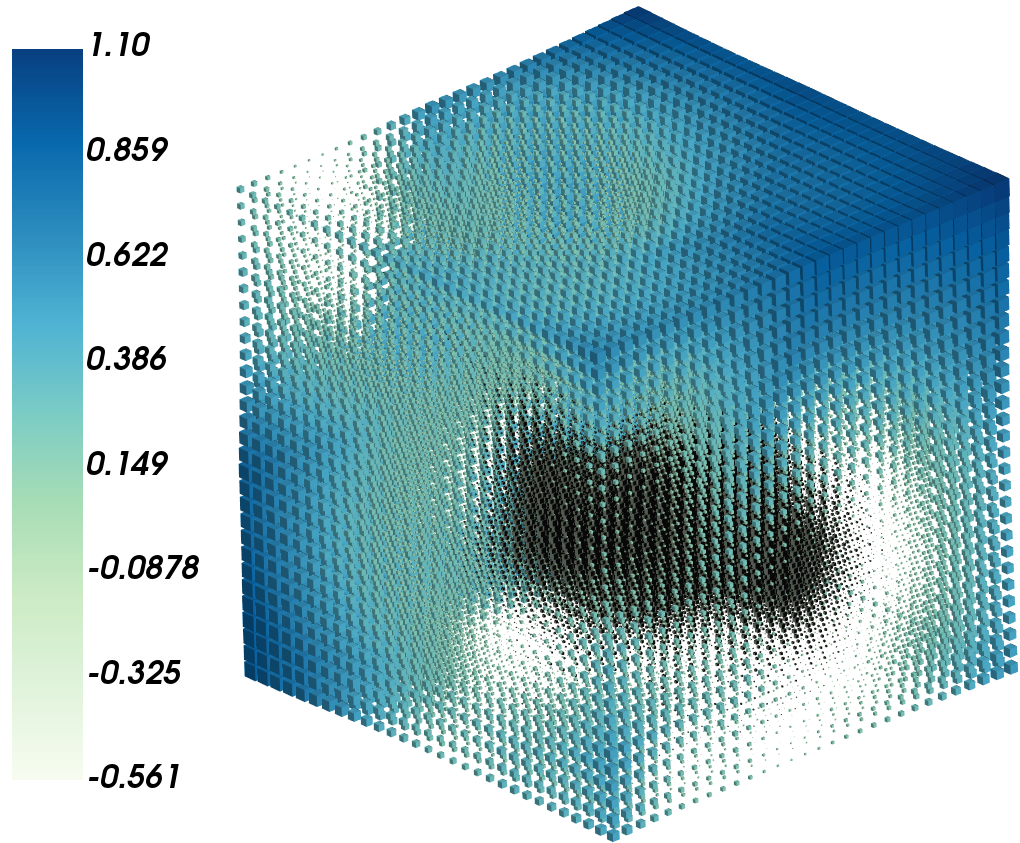}
    \def\cc{30mm}
    {\footnotesize
    (a) Explicit representations \hspace{20mm}
    (b) Voxels \hspace{\cc}
    (c) Point cloud \hspace{\cc}
    (d) Level set }
    \caption{Four common representations of 3D shape along with  
    their advantages and disadvantages. }
    \label{fig:representations_bunny}
\end{figure*}

Each of these representations 
come with their own advantages and disadvantages, in particular for 
the application of shape inference in a learning framework, figure \ref{fig:representations_bunny}. 
Explicit representations, such as triangle meshes are exceedingly popular in the graphics community as they 
provide a compact representation able to capture detailed geometry of most 3D objects. However, they are irregular in nature, not uniquely defined, and they cannot be easily integrated into learning frameworks. 
Voxel occupancy maps on the other hand are defined on fixed regular grids making them exceptionally well suited for learning applications, in particular 
convolutional approaches. However, unless the resolution of the tessellated grid 
is high this class of representations typically result in coarse reconstructions. 
Point clouds are also commonly used to describe the shape of 3D objects. 
However, this approach suffers from many of the same drawbacks as polygon meshes 
and is, in addition, only able to provide sparse representations of shapes. 
In this work we instead argue that implicit representations, or level sets, 
constitutes a more appropriate choice for the task of learned shape 
inference. Similar to voxels, level sets are defined on regular grids, making them directly suitable for the use with convolutional neural networks.
However, this formulation is also more expressive and able to capture more 
geometrical information of 3D shapes resulting in higher quality inferences.
Furthermore, level sets are also equipped with a very elegant mathematical formulation 
that permits the inclusion of additional geometric quantities,
such as surface orientation, smoothness and volume, in a very elegant manner. 
To the best of our knowledge such a direct level set formulation and its geometrical 
properties have not yet been exploited in previous works. 

Convolutional neural networks~\cite{liu2015deep,girdhar2016learning,dosovitskiy2015learning,rezende2016unsupervised,hu2017deep} and generative adversarial models (GANs)\cite{choy20163d,niessner2013real} have been successfully applied to 3D reconstruction problems by using either volumetric or point cloud representations. The success is mainly due to the availability of large-scale datasets of 3D objects such as Shapenet \cite{chang2015shapenet} and ObjectNet3D \cite{xiang2016objectnet3d}.

All aforementioned approaches require additional step of applying meshing techniques such as SSD or marching cubes to extract the actual 3D mesh. More specifically, one of the main limitation of the existing deep learning approaches for 3D reconstruction is that they are unable to classify pixels lying on the boundary of the object accurately. Thus, the generated boundaries are fuzzy and inaccurate resulting in a coarse and discrete representation of 3-dimensional object. This is specifically due to the fact that a more efficient representations of 3D objects such as polygon mesh do not fit well to deep neural architectures and poses problems in performing back propagation.

In the light of above discussion, we propose to generate a continuous representation of the reconstructed object by integrating level set methods in deep convolutional neural networks. The level set method introduced in \cite{dervieux1980finite,osher1988fronts}, and successfully applied in segmentation and medical image analysis \cite{whitaker1998level,kichenassamy1995gradient}, is a mathematically elegant way of implicitly representing shape of an object and its boundary evolution in time, which can be represented as a zero level set of an embedding function. To the best of our knowledge, incorporating a level set methods in a deep end-to-end trainable model and representing the 3D output as a level set of a continuous embedding function has never been studied in the literature.

We demonstrate that incorporating level set representation in an end-to-end trainable network can lead to a more accurate reconstruction. To evaluate this, we used the ShapeNet dataset along with its labeled subset ShapeNet-Core, and compared our approach against voxel-based shape representation. 
We deliberately chose a simple deep architecture which encodes 3-dimensional objects into 64-dimensional vectors and decodes that representation back into the 3-dimensional object. 
As evidenced in the experiments, our reconstruction is much more accurate than that of using voxel representations, clearly showing that the improvement in representation is due to the level set incorporation, rather than to complex deep architectures. Moreover, representing the output as a level set of a continuous embedding function enables our model to introduce various regularisers, giving an advantage over classical volumetric methods.

\section{Related Work}
\label{sec:related}

3D reconstruction is a fundamental problem in computer vision with many potential applications such as robotic manipulation, self-driving cars, and augmented reality. Existing 3D reconstruction methods can be divided into two broad categories: reconstruction from a single image~\cite{criminisi2000single}, and from multiple images (\eg structure from motion~\cite{haming2010structure}). 

One of the important challenges in stepping towards solving this problem is the limited access to the large amount of data required for an accurate reconstruction. Recently, large-scale datasets of 3D objects such as ShapeNet~\cite{chang2015shapenet} and ObjectNet3D~\cite{xiang2016objectnet3d} have been made available which allowed the field to make great progress. There have also been attempts on using prior knowledge about the shape of 3D objects~\cite{dame2013dense} in the absence of large amounts of data. Despite its effectiveness, the described approaches relies on hand-crafted features which limits its scalability.

With the advent of deep learning architectures, convolutional neural networks have found to be very
useful in 3D reconstruction using only a single image ~\cite{liu2015deep}. Recently, \cite{girdhar2016learning} and ~\cite{dosovitskiy2015learning} proposed the use of shape and camera features along with the images, respectively. Despite their success, these methods rely on ground truth which is not a realistic scenario.

To tackle this problem, different CNNs-based approaches have been introduced which require only weak supervision ~\cite{yan2016perspective,rezende2016unsupervised}, and they are able to handle more shape variations. However, they do not scale well when increasing the resolution of the input image. Moreover, more efficient representations of 3D objects like polygon meshes do not easily fit into DNNs architectures.

Recurrent neural networks have recently been proposed to infer 3D shapes. \cite{choy20163d} introduced generative adversarial models (GANs) using long short-term memory (LSTM) for reconstructing voxels or point clouds achieving state-of-the-art results. \cite{rezende2016unsupervised} proposed the use of conditional GANs in an unsupervised setting and \cite{steinbrucker2014volumetric} proposed the use of octrees. An important drawback of GAN-based methods is that they are computationally expensive and not accurate when using metrics such as the Chamfer distance, Earth Mover's distance or intersection over union (IoU). Another drawback of such methods is that they do not allow multiple reconstruction which is sometimes needed when dealing with single image reconstruction. As a response to these shortcomings Delaunay Tetrahydration or voxel block hashing ~\cite{niessner2013real} were introduced.

Even though great progress has been achieved in the 3D reconstruction field, the aforementioned approaches suffer from the lack of geometry due to its poor shape representation. In this paper, we propose the use of a continuous 3D shape representation by integrating level sets into CNNs. Our aim is to infer embedding functions to represent the geometry of a 3D shape where we can then extract its level set to have a continuous shape representation, \ie a 3D surface.

\section{Preliminaries}
\label{sec:background}

\textbf{Level Set Surface Representations.}
The Level Set method for representing moving interfaces was proposed independently 
by \cite{osher1988fronts} and \cite{dervieux1980finite}. 
This method defines a time dependent orientable surface $\Gamma(t)$ implicitly as the zero isocontour, or level set, of a higher dimensional 
auxiliary scalar function, called the \emph{level set function}
or \emph{embedding function}, $\phi(x,t): \Omega \times \R \mapsto \R$,  as, 
\begin{align} 
\Gamma(t)= \left\{ x : \phi(x,t) = 0 \right\},
\end{align}
with the convention that $\phi(x,t)$ is positive on the interior and 
negative on the exterior of $\Gamma$. 
The underlying idea of the level set method is then to capture the motion of 
the isosurface through the manipulation of the level set function $\phi$.

Given a surface velocity $v$, the evolution of the isosurface $\Gamma$ 
is particularly simple, it is obtained as the solution 
of the partial differential equation (PDE) (known as the  \emph{level set equation}) 
\begin{align}
\frac{\partial \phi}{\partial t} = v |\nabla \phi|. 
\end{align}
In practice, this problem is discretised and numerical computations 
are performed  on a fixed Cartesian grid in some domain.
This formulation also permits a natural way to calculate additional interface primitives, 
\ie surface normals, curvatures and volumes. 
Such primitives are typically used in applications involving entities with physical meanings, 
to impose specific variabilities of the obtained solution, for instance to favour smoothness of the surface $\Gamma$.

One additional advantage of the level set formulation is that it allows 
complex topologies as well as changes in topology in a very elegant 
and simple manner without the need for explicit modelling. 
This is typically not the case in most parametric approaches, where 
topological variations needs to be handled explicitly through dedicated 
procedures.

\textbf{Minimal Oriented Surface Models.}
Here we formulate the task of 
fitting an implicitly defined closed surface $\Gamma$ to a given oriented surface $\S \subset \R^3$ as that of simultaneously minimising the distance to a discrete number of points $x_i \in \S$ as well as the difference between the orientation of the unit-length surface normals $n_i$ (at $x_i$) and the normals of $\Gamma$. 
Note that $\S$ does not necessarily have to be a closed surface, hence the 
orientation of the normals $n_i$ are not uniquely defined and only determined up to a sign 
ambiguity (\ie $n_i \sim \pm n_i$).  
Let $\S$ be given as a  
collection of $m$ data points of $\X =\{x_i\}_{i=1}^m$ and their corresponding normals 
$\N =\{n_i\}_{i=1}^m$, and let $d_\X(x)$ denote as the distance 
distance function to $\X$,  
\begin{align}
d(x,\X) = \inf_{y\in \X} \|x-y\|. 
\end{align}
As in \cite{zhao2000implicit}, we then define the following energy functional for the 
variational formulation,  
\begin{align}
    E_\X(\Gamma)= \left( \int_\Gamma d(s, \X)^p  ds
    \right)^{1/p}
    , 1\leq p \leq \infty. 
\end{align}
The above functional measures the deviation as the $L_p$-norm of the distance from the surface $\Gamma$ from the point set $\X$.

Similarly, for the normals $\N$ we define an energy functional that quantifies the 
difference between the normal of the estimated surface $\Gamma$ and the desired 
surface normals of the given surface $\S$. The measure we propose is the 
$L_p$-norm of the angular distance between the normals of $\Gamma$  and those of $\N$. 
\begin{align}
    E_\N(\Gamma)= \left( \int_\Gamma 
\left(1 - \left|N(s)\cdot n_\Gamma(s) \right| \right)^{p}  ds
    \right)^{1/p},  1\leq p \leq \infty, 
    \label{eq:EX}
\end{align}
where $N(s) = n_i$ when $x_i$ is the closest point to $s$. 
With the outward unit normal of $\Gamma$ given by 
\begin{align}
n_\Gamma(s)=\frac{\nabla \phi(s)}{\|\nabla \phi(s)\| },
\end{align}
we can write $E_\N(\Gamma)$ as 
\begin{align}
    E_\N(\Gamma)= \left( \int_\Gamma 
\left(1 - \left|N(s)\cdot \frac{\nabla \phi(s)}{\|\nabla \phi(s)\| } \right| \right)^{p}  ds
    \right)^{1/p}.
    \label{eq:EN}
\end{align}
Note that since both \eqref{eq:EX} and \eqref{eq:EN} are defined as 
surface integral over $\Gamma$ they will return decreased energies on 
surfaces with smaller area. 
Consequently, both these energy functionals contain an implicit smoothing 
component due to this predilection towards reduced surface areas.

\textbf{Shape Priors \& Regularisation.}
Attempting to impose prior knowledge on shapes can be a very useful proposition in a wide range of applications. 
A distinction is typically made between generic (or geometric) priors and object specific priors. 
The former concerns geometric quantities, generic to all shapes, such as \emph{surface area}, 
\emph{volume} or \emph{surface smoothness}. 
In the latter case, the priors are computed from set of given samples of a specific object 
of interest. Formulations for incorporating such priors in to the level set framework has 
been the topic of considerable research efforts, for an excellent review see \cite{cremers2007review}. 

For the sake of simplicity and brevity, in this section we limit ourselves to two 
of the most fundamental generic shape priors, \emph{surface area} and 
\emph{volume}. 
They are defined as, 
\begin{align}
E_{area}= \int_\Gamma ds,
\end{align}
and
\begin{align}
E_{vol}= \int_{\textnormal{int} \Gamma} ds. 
\end{align}
However, many of the additional shape priors available can presumably be 
directly incorporated in to our proposed framework as well. 

\textbf{Embedding functions and Ill-Conditioning.}
It has been observed that in its conventional formulation the level set function often develop  
complications related to ill-conditioning during the evolution process, \cite{gomes2000reconciling}. 
These complications may in turn lead to numerical issues and result in an unstable surface motion. 
Many of these conditioning issues are related to degraded level set functions, ones that are either 
too steep or too flat near its zero level set. A class of functions that do not display these 
properties are the \emph{signed distance functions}. They are defined as 
\begin{align}
f(x) = \pm \inf_{y\in \Gamma} \|x-y\|, 
\end{align}
where $f(x)$ is $>0$ if $x$ is in the interior of $\Gamma$ and negative otherwise. 
Signed distance functions have unit gradient, $|\nabla f|=1$, not only 
in the proximity to $\Gamma$ by its entire domain. 
Consequently, a common approach to overcoming these stability issues is to regularly correct or 
\emph{reinitialise} the level set function to be the signed distance function of the current 
zero level set isosurface. 

However, in our intended setting of shape inference in a learning framework, such a
reinitialisation procedure is not directly applicable. Instead we propose the use of an 
energy functional, similar to the work of \cite{li2010distanceregularized}, that promotes the unit gradient property, 
\begin{align}
    E_{sdf}(\phi)=  \int (\|\nabla \phi(x)\|-1)^2 dx.
\end{align}

\section{A Variational Loss Function for Shape Inference}
\label{sec:algorithm}
In this section we show how an implicit representation of 3D surfaces
can be introduced to a learning framework through a direct application of 
the variational formulations of the previous section.

Given a set of $n$ training examples $I^j$ and their corresponding 
ground truth oriented shapes $\S^j=\{\X^j, \N^j\}$, here represented 
as a collection of discrete points with associated normals, see section  
\ref{sec:background}. 
Let $\theta$ denote the parameters of some predictive procedure, a neural network, 
that from an input $I$ estimates shape implicitly through a level set function,
$\tilde{\phi}(I;\theta)$. 
At training, we then seek to minimise (with respect to $\theta$) the dissimilarity (measured by a loss function) between the training data and the predictions made by our network. 
The general variational loss function we propose in this work is as follows, 
\begin{align}
&    L(\theta) = \sum_{j\in \D} E_{\X^j}(\Gammat(I^j;\theta)) 
    + \alpha_1 \sum_{j\in \D} E_{\N^j}(\Gammat(I^j;\theta))    \nonumber \\ 
&    +\alpha_2 \sum_{j\in \D} E_{sdf}(\tilde{\phi}(I^j;\theta))
+ \alpha_3 \sum_{j\in \D} E_{area}(\Gammat(I^j;\theta))   \nonumber \\ 
&+ \alpha_4 \sum_{j\in \D} E_{vol}(\Gammat(I^j;\theta)).
\label{eq:lossfunction}
\end{align}
Here $\Gammat$ denotes the zero level set of the predicted level set function $\tilde{\phi}$ given input $I$, that is $\Gammat(I;\theta)=\{ x \ : \ \tilde{\phi}(I;\theta)=0 \}$, $\D=\{1,...,n\}$ and $\alpha_1-\alpha_4$ are 
weighting parameters.

By introducing the Dirac delta function $\delta$ and the Heaviside function $H$
we can write the individual components of \eqref{eq:lossfunction} as, 
{\allowdisplaybreaks
\begin{align}
&\sum_{j\in \D} E_{\X^j}(\Gammat(I^j;\theta)) = \nonumber \\
& \hspace{10mm}=\sum_{j\in \D} \left( \int_{\R^3} \delta(\phit(x,I^j;\theta)) d(x, \X^j)^p  dx \right)^{1/p},
\label{eq:losscomponents1}
\\
&\sum_{j\in \D} E_{\N^j}(\Gammat(I^j;\theta))  = 
\sum_{j\in \D} 
\Bigg( \int_{\R^3} \delta(\phit(x,I^j;\theta))
\nonumber \\
& \hspace{11mm} \Big(1 - \Big|N^j(x)\cdot \frac{\nabla \phit(x;I^j,\theta)}{\|\nabla \phit(x,I^j;\theta)\| } \Big| \Big)^{p}  dx \Bigg)^{1/p},
\label{eq:losscomponents2}
\\ 
&\sum_{j\in \D} E_{sdf}(\phit(I^j;\theta)) = 
\sum_{j\in \D} \int_{\R^3} (\|\nabla \phit(x,I^j;\theta)\|-1)^2 dx, 
\label{eq:losscomponents3}
\\
&\sum_{j\in \D} E_{area}(\Gammat(\theta,I^j)) = 
\sum_{j\in \D} \int_{\R^3} \delta(\phit(x,I^j;\theta)) \ dx, 
\label{eq:losscomponents4}
\\
&\sum_{j\in \D} E_{vol}(\Gammat(\theta,I^j)) = 
\sum_{j\in \D} \int_{\R^3} H(\phit(x,I^j;\theta)) \ dx.
\label{eq:losscomponents5}
\end{align}
}
In practise the above loss function is only evaluated on a fixed equidistant grid $\Omega$
in the volume of interest. It is then also necessary to introduce continuous 
approximations of the Dirac delta function and Heaviside function, 
Following the work of  \cite{zhao1996variational} we use the following 
$C^1$ and $C^2$ approximations of  $\delta$ and $H$ respectively, 
\begin{align}
    \delta_\epsilon(x)=\left\{
    \begin{array}{cc}
    \frac{1}{2\epsilon}\left( 1 + \cos (\frac{\pi x}{\epsilon}) \right), &  |x|\leq \epsilon,\\
         0, & |x|> \epsilon,
    \end{array}
    \right.
    \label{eq:diracapprox}
\end{align}
and
\begin{align}
    H_\epsilon(x)=\left\{
    \begin{array}{cc}
         \frac{1}{2}\left( 1+\frac{x}{\epsilon}+\frac{1}{\pi}\sin(\frac{\pi x}{\epsilon}) \right), &  |x|\leq \epsilon,\\
         1, & x> \epsilon,\\
         0, & x> -\epsilon,
    \end{array}
    \right.
    \label{eq:heavisidecapprox}
\end{align}
note that here $H_\epsilon'(x)=\delta_\epsilon(x)$. 
Inserting \eqref{eq:diracapprox}-\eqref{eq:heavisidecapprox} in 
\eqref{eq:losscomponents1}-\eqref{eq:losscomponents5} we obtain 
an approximated loss function $L_\epsilon$ expressed entirely in $\phit$.
With the simplified notation $\phit^j(x)=\phit(x,I^j;\theta))$ and $d^j(x)^p=d(x, \X^j)^p$,
we arrive at, 
\begin{align}
&    L_\epsilon(\theta) = 
    \sum_{j\in \D} \left( \sum_{x\in\Omega} \delta_\epsilon(\phit^j(x)) d^j(x)^p \right)^{1/p}  \\ \nonumber
&    + \alpha_1     \sum_{j\in \D} \Bigg( \sum_{x\in\Omega} 
    \delta_\epsilon(\phit^j(x))
 \Big(1 - \Big|N^j(x)\cdot \frac{\nabla \phit^j(x)}{\|\nabla \phit^j(x)\| } \Big| \Big)^{p}   \Bigg)^{1/p}\\ \nonumber
&    +\alpha_2 \sum_{j\in \D} \sum_{x\in\Omega} (\|\nabla \phit^j(x)\|-1)^2 
+ \alpha_3 \sum_{j\in \D} \sum_{x\in \Omega} \delta_\epsilon(\phit^j(x))\\ \nonumber
&+ \alpha_4 \sum_{j\in \D} \sum_{x\in \Omega} H_\epsilon(\phit^j(x)).
\label{eq:lossfunctioneps}
\end{align}
Finally, differentiating $L_\epsilon$ with respect to $\phit$ on the discrete grid $\Omega$
yields,
{\allowdisplaybreaks
\begin{align}
    &\frac{\partial L_\epsilon}{\partial \phit}= \nonumber \\ \nonumber
    &=\sum_{j\in \D} \frac{1}{p}\Big( \sum_{x\in \Omega} \delta_\epsilon(\phit^j(x)) d^j(x)^p   \Big)^{\frac{1-p}{p}} \delta_\epsilon'(\phit^j(x)) d^j(x)^p 
     \\ \nonumber
    %
    & +\frac{\alpha_1}{p}
 	\sum_{j\in \D} 
 	\Bigg( \sum_{x\in\Omega} 
    \delta_\epsilon(\phit^j(x))
 \Big(1 - \Big| \frac{N^j(x)\cdot\nabla \phit^j(x)}{\|\nabla \phit^j(x)\| } \Big| \Big)^{p}   \Bigg)^{\frac{1-p}{p}}    \\ \nonumber
&
\Bigg( \delta'_\epsilon(\phit^j(x))
 \Big(1 - \Big| \frac{N^j(x)\cdot\nabla \phit^j(x)}{\|\nabla \phit^j(x)\| } \Big| \Big)^{p}  
+ \nonumber \\
&     \delta_\epsilon(\phit^j(x))
 \frac{\partial }{\partial \phit} 
 \Big(1 - \Big| \frac{N^j(x)\cdot\nabla \phit^j(x)}{\|\nabla \phit^j(x)\| } \Big| \Big)^{p}  
\Bigg) 
%
     \\ \nonumber
    &+\alpha_2 
    \sum_{j\in \D} \sum_{x \in \Omega} 
    (\|\nabla \phit^j(x)\|-1) \ \nabla \cdot \left( \frac{\nabla \phit^j(x)}{||\nabla \phit^j(x)||}\right)
     \\ \nonumber
&+ \alpha_3 \sum_{j\in \D} \delta_\epsilon'(\phit^j(x)) 
+ \alpha_4 \sum_{j\in \D}  \delta_\epsilon(\phit^j(x)).
\end{align}
}
Evaluating $L_\epsilon$ and $\frac{\partial L_\epsilon}{\partial \phit}$, 
for a given level set function $\phit$, is entirely straightforward, only 
requiring 
the calculation of the inferred level set $\Gammat$ from $\phit$. This can be done 
using any of a number of existing algorithms, see \cite{hansen2011visualization}. 
Note that, as a consequence, our proposed framework is entirely indifferent to the 
choice isosurface extraction algorithm used. This is an important distinction from 
work such as 
\cite{liao2018deep} which is derived from a very specific choice of algorithm.

\section{Experimental Validation}
\label{sec:experiments}

\begin{table*}[t!]
\centering
\ra{1.3}
\caption{Performance comparison between voxel occupancy and level set representations on 
training / test data with two different resolutions, $20^3$ and $30^3$, measured by IoU (in \%). 
Here $\Delta$ denotes the difference in IoU on the test data.
\label{tab:IoU}} \vspace{1ex}
\begin{tabular}{@{}lrrrcrrrcrrrr@{}}
\toprule
\textbf{IoU [\%]}&& \multicolumn{3}{c}{$20^3$} & \phantom{a}& \multicolumn{3}{c}{$30^3$} \\
\cmidrule{3-5} \cmidrule{7-9} 
Category& \phantom{ab} & voxels & $\phi$ & $\Delta$ & & voxels & $\phi$ & $\Delta$
\\ \midrule
Bottle && 73.5 / 64.9  & 82.2 / \textbf{78.4}  &(+13.5)&& 84.9 / 72.9  &  87.1 / \textbf{82.8} &(+9.9)  \\ 
Car && 76.2 / 74.7  &  88.1 / \textbf{86.6}    & (+11.9)&& 86.8 / 81.5 &  88.3 / \textbf{86.9} &(+05.4)   \\ 
Chair && 75.1 / 57.8  &  76.5 / \textbf{64.0}  &(+6.2)&&  86.1 / 60.0  & 77.8 / \textbf{62.5} &(+02.5)   \\ 
Sofa && 75.8 / 62.7  &  75.3 / \textbf{68.9}  &(+6.2)&& 87.1 / 68.6  & 80.2 /   \textbf{73.0} &(+04.4)   \\ 
\bottomrule
\end{tabular}

\vspace{5mm}
\centering
\ra{1.3}
\caption{Performance  comparison between voxel occupancy and level set representations on training/test 
data with two different resolutions, $20^3$ and $30^3$, measured by the Chamfer 
distance.
Here $\Delta$ denotes the difference in Chamfer distance on the test data.
\label{tab:chamfer}} \vspace{1ex}
\begin{tabular}{@{}lrrrcrrrcrrrr@{}}
\toprule
\textbf{Chamfer}&& \multicolumn{3}{c}{$20^3$} & \phantom{a}& \multicolumn{3}{c}{$30^3$} \\
\cmidrule{3-5} \cmidrule{7-9} 
Category& \phantom{ab} & voxels & $\phi$ & $\Delta$ & & voxels & $\phi$ & $\Delta$
\\ \midrule
    Bottle && 0.0830 / 0.0949 &  0.0593 / \textbf{0.0652}  &(-0.0297)&& 0.0579 / 0.0675 &  0.0440 / \textbf{0.0447} &  (-0.0228) \\ 
    Car && 0.0899 / 0.0874  &  0.0386 / \textbf{0.0411}  &(-0.0463)&& 0.0633 / 0.0658   &  0.0369 / \textbf{0.0390}   &  (-0.0268)   \\ 
    Chair && 0.0833 / 0.0970 &  0.0610 / \textbf{0.0855}  &(-0.0115)&& 0.0584 / \textbf{0.0827} &  0.0591 / 0.0869 &    (+0.0042)\\ 
    Sofa && 0.0823 / 0.0935  &  0.0520 / \textbf{0.0649}  &(-0.0286)&&  0.0576 / 0.0733  & 0.0471 / \textbf{0.0595} &   (-0.0138) \\ 
\bottomrule
\end{tabular}
\end{table*}

In this section we present our empirical evaluation of the proposed 
variational formulation for 3D shape inference from single 2D images. 
These experiments were primarily directed at investigating the 
potential improvement obtained by an implicit representation  
over that of more conventional representation. 
This paper was not intended to study the suitability of different 
types of networks for the task of shape inference. 
In fact, as discussed further down in this section, 
we deliberately chose a rather simple network architecture to conduct 
this study on. 

\subsection{Implementation Details}
We begin by discussing some of the practical aspects of the experimental setup we 
used in this section.  

\textbf{Dataset \& Preprocessing.}
We evaluated the proposed formulation on data from the ShapeNet dataset \cite{chang2015shapenet}. We chose a subset of $4$ categories 
from this dataset: \emph{'bottles'}, \emph{'cars'}, \emph{'chairs'} and \emph{'sofas'}.
As ShapeNet models often do not have an empty interior, we used the manifold 
surface generation method of \cite{huang2018robust} as a preprocessing stage 
to generate closed manifolds of those models and used them as ground truth.

We ended up with approximately $500$ models for \emph{'bottles'} and 
$2000$ models each for 
the categories \emph{'cars'}, \emph{'chairs'} and \emph{'sofas'}. 
Each model is rendered into $20$ 2D 
views, (input images) using fixed elevation, and equally spaced 
azimuth angles. This data was then randomly divided into  $80/20$ train-test 
splits. The ground-truth manifolds are also converted to a voxel occupancy map, 
for training and testing the voxel-based loss functions, 
using the procedure of \cite{nooruddin2003simplification}.

\begin{figure}
    \centering
    \includegraphics[width=\linewidth]{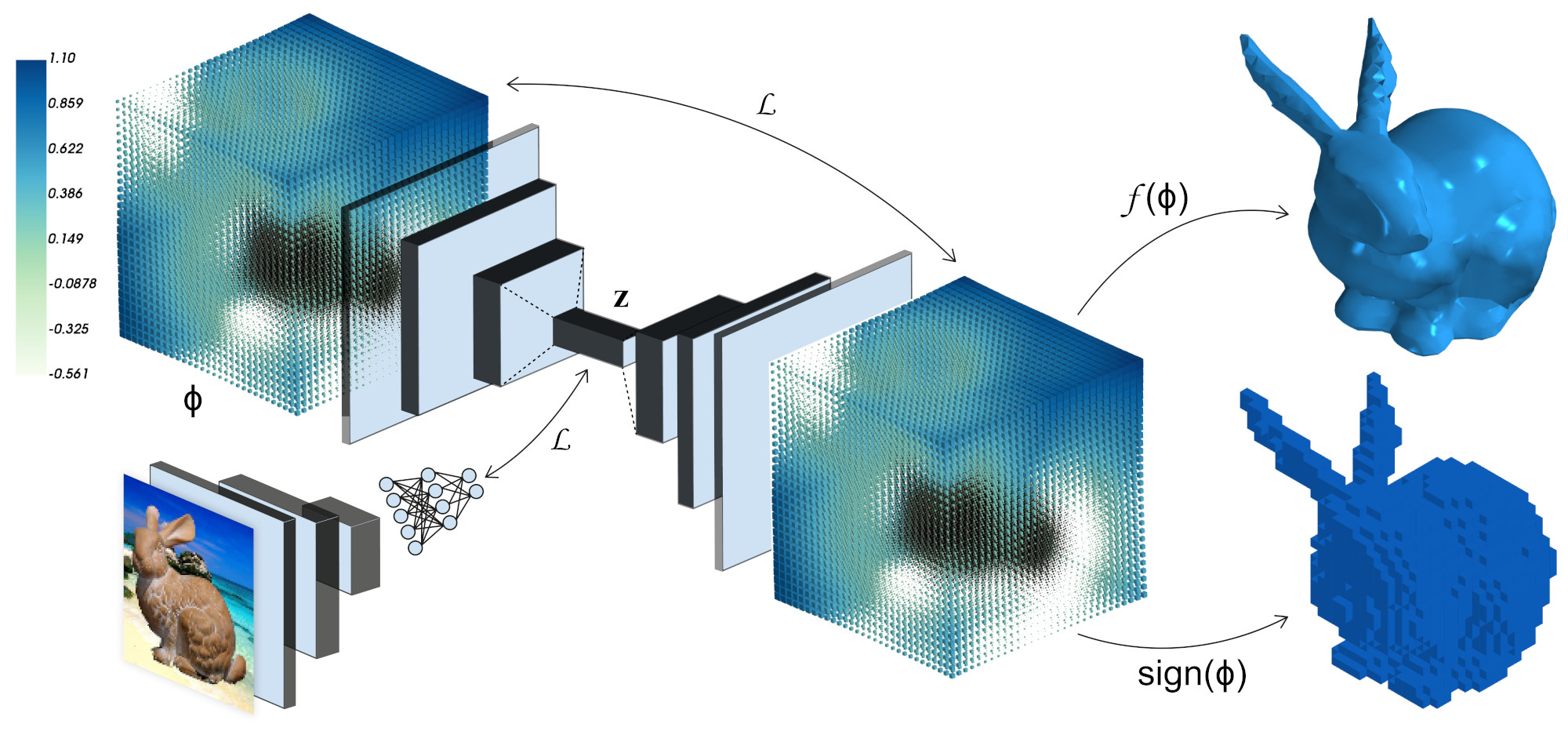}
    \caption{An overview of the network architecture used.}
    \label{fig:architecture}
\end{figure}

\textbf{Network Architecture.}
Motivated by~\cite{girdhar2016learning}, we use a simple 3D auto-encoder network which predicts 3D representation from 2D rendered image, and consists of two components: an auto-encoder as a generator and a CNN as a predictor connected by a 64-dimensional vector embedding space. 

Specifically, the autoencoder network with convolution and deconvolution layers, projects a 3D shape to the 64-dimensional space, and decodes it back to a 3D shape. The encoder composed of four convolutional layers and a fully connected layer to project the data to the 64D embedding vector. The decoder consists of five 3D convolutional layers with stride 1 connected by batch norm and 
ReLU non-linearities to map the  embedding vector back to 3D space.

Similar to MobileNetV2 \cite{sandler2018mobilenetv2} architecture, the CNN comprised of five convolutional layers, two fully connected layers, and an added 64 dimensional layer, initialised with the ImageNet \cite{deng2009imagenet} weights, and projects a 2D image to the 64 dimensional space.

The two components are jointly optimised at training time, taking the 3D CAD model along with its 2D image. At test time, the encoder part of the auto-encoder is removed, and then the ConvNet and the decoder are used to obtain a 3D representation and images in the shared latent space.

Note that the reason behind our choice of such a simple architecture is to demonstrate that the improvement in the representation is due to our representation of shape, rather than 
adopting complex deep architectures.

\textbf{Comparison.}
We compare the proposed implicit shape representation 
to that of a voxel-based occupancy map representation. 
This is done by training the network using two distinct loss functions: 
the variational loss function, defined in section \ref{sec:algorithm}
(with $p=2$, $\epsilon=0.15$, $\alpha_1=0.8$, $\alpha_2=1$ and 
$\alpha_3=\alpha_4=0.1$), 
and the voxel-based cross-entropy loss of \cite{girdhar2016learning}, 
\begin{align}
	E(\hat{p})=-\frac{1}{N}\sum_{n=1}^N [p_n \log \hat{p}_n +( 1-p_n) \log (1-\hat{p}_n)],
\end{align}
where $p$ denotes the ground-truth voxel occupancy, $N$ the total number of voxels 
and $\hat{p}$ the prediction. 
Both formulations were trained with 2D images as inputs, for 3000 epochs using a  
batch size of 64 and a learning rate of $10^{-6}$,  
the ground-truth shapes are represented as manifolds and voxel occupancy maps respectively. 
It is important to note that the architecture is exactly identical in 
both instances, consequently so is the memory requirements and computational 
costs at evaluation. The only slight difference is that the ground truth for 
the voxel-based approach is binary  as opposed to real-valued 
(a polygon mesh) for the variational formulation. 

We deliberately limited our trials to the above two formulations only. 
This was mainly motivated by the fact that it permitted us to use the same network architecture throughout the entire experiment, thus resulting in a very fair 
comparison. 
In addition, our proposed formulation should not be viewed as a competitor to 
existing algorithms  but rather as a complement. 
The use of implicit shape representations could readily be incorporated 
into many of the existing approaches currently in use. 

\textbf{Evaluation Metrics.}
Here we considered two separate metrics for evaluating shapes inferred by our 
network,  the \emph{Intersection over Union} (IoU), also known as the \emph{Jaccard Index} 
\cite{levandowsky1971distance}, and the Chamfer distance \cite{barrow1977parametric}. 

The IoU measures similarity between two finite sample sets, $A$ and $B$. 
It is defined as the ratio between the size of their intersection and union, 
\begin{align}
    \IoU(A,B)=\frac{|A \cap B|}{|A \cup B|}.
\end{align}
This metric ranges between $0$ and $1$, with small values indicating low 
degrees of similarity and large values indicating high degrees of similarity. 

The Chamfer distance is a symmetric metric, here used to measure distance between 
a pair of surfaces, or more accurately between two finite point clouds ($\P_1$ and $\P_2$) 
sampled from a pair of surfaces. 
\begin{align}
    d_{\textnormal{ch}}(\P_1,\P_2)= 
    \frac{1}{|\P_1|} &\sum_{x\in\P_1} \min_{y\in\P_2} ||x-y|| \nonumber \\
    &+\frac{1}{|\P_2|} \sum_{y\in\P_2} \min_{x\in\P_1}  ||y-x||.
\end{align}
Note that these two measures are defined on distinctly different domains, sets and 
point-clouds/surfaces for IoU and Chamfer distance respectively. 
As we are comparing different shape representations, the ground truth and level set 
representations are available as surfaces and the voxel occupancies as sets, 
we need to be attentive to how these measures are applied. 

The IoU is measured by first applying a voxelization procedure  
\cite{nooruddin2003simplification} to the ground truth as well as the level sets inferred 
by the trained network. To ensure that all the finer details of these surfaces are preserved, 
this voxelization should be performed at a high resolution.\footnote{We empirically observed that 
going beyond a resolution of $128^3$ did not impact the results noticeably.} 
Correspondingly, surface representations can be extracted from occupancy grids by means of 
the Marching Cubes algorithm \cite{lorensen1987marching}.

\subsection{Results}

We evaluated the efficiency of our proposed variational-based implicit shape representation 
both quantitatively and qualitatively. 
Quantitatively, we measured the similarity/distance 
between the ground-truth and the inferred shapes using both a voxel and a level set 
representation. 
The results are shown in tables \ref{tab:IoU} and \ref{tab:chamfer}.

These results appear very promising and supports our argument 
that in learning frameworks implicit shape representations are able to 
express shape more accurately and with more detail than voxel based representations can. We can further see that, as expected, the difference between these two representations is reduced as the resolution increases. 
Voxels does appear to perform on par with, or even slightly better than level sets in one instance, on the class 'chair'. 
We believe this can be explained by the topology of that specific class. Many chairs typically have very long thin legs, structures that are difficult to capture at a low resolution,  both for voxels and level sets. 

\begin{figure*}[p]
    \def\dd{.135}
    \centering
       {\fontfamily{ptm}\selectfont {\scriptsize \textbf{\hspace{0pt} (a) GT \hspace{35pt} (b) Input image \hspace{27pt} (c) Level set $20^3$ \hspace{22pt} (d) Voxel $20^3$ \hspace{22pt} (e) Level set $30^3$ \hspace{25pt} (f) Voxel $30^3$}}} \\
    \includegraphics[width=\dd\textwidth]{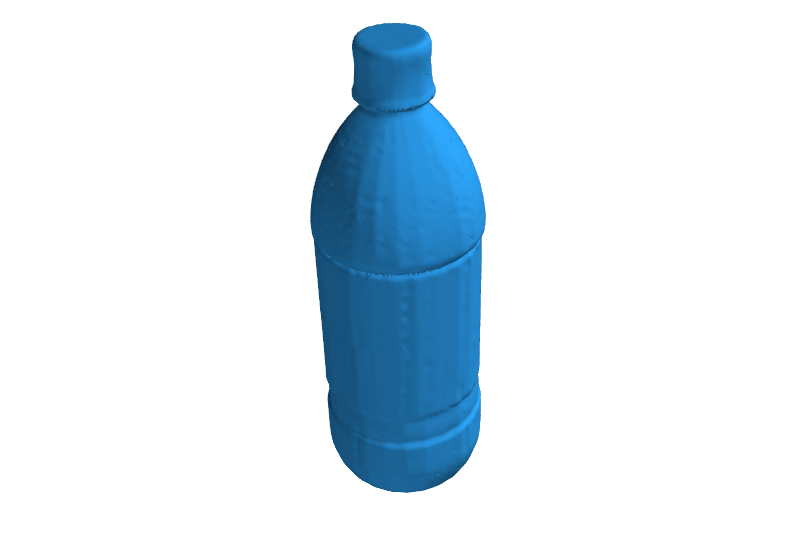}\hspace{1mm} 
    \includegraphics[width=\dd\textwidth]{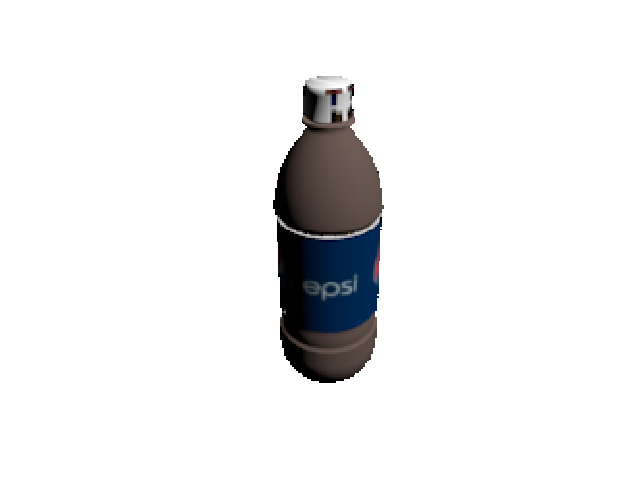}\hspace{1mm} 
    \includegraphics[width=\dd\textwidth]{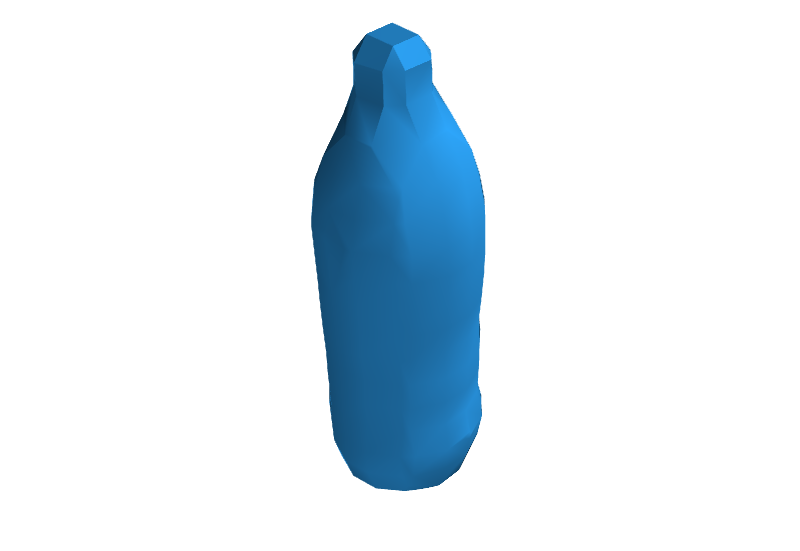}\hspace{1mm}
    \includegraphics[width=\dd\textwidth]{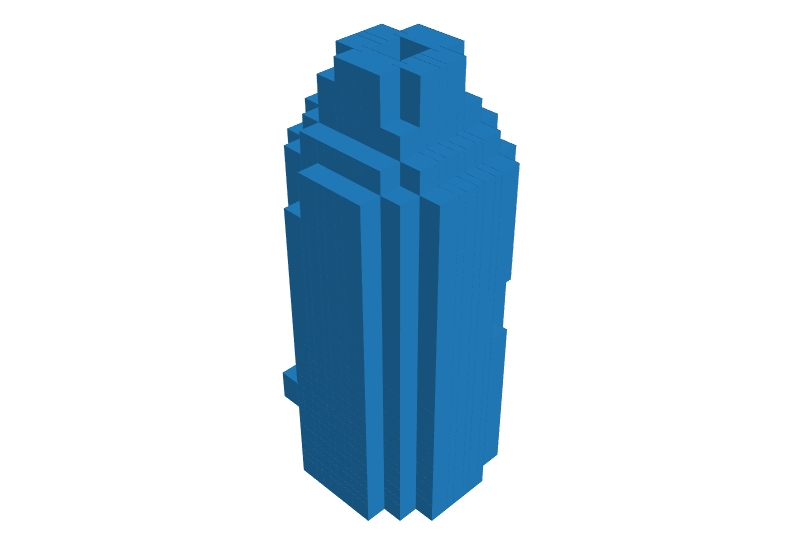}\hspace{1mm}
    \includegraphics[width=\dd\textwidth]{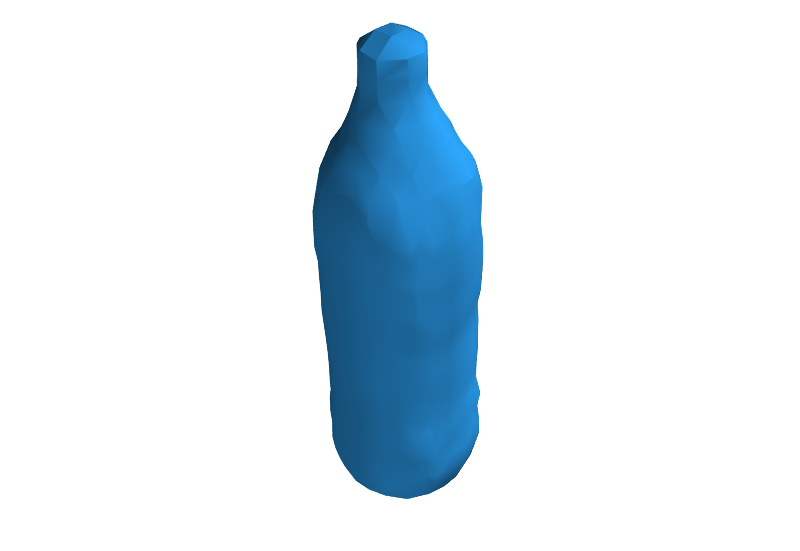}\hspace{1mm}
    \includegraphics[width=\dd\textwidth]{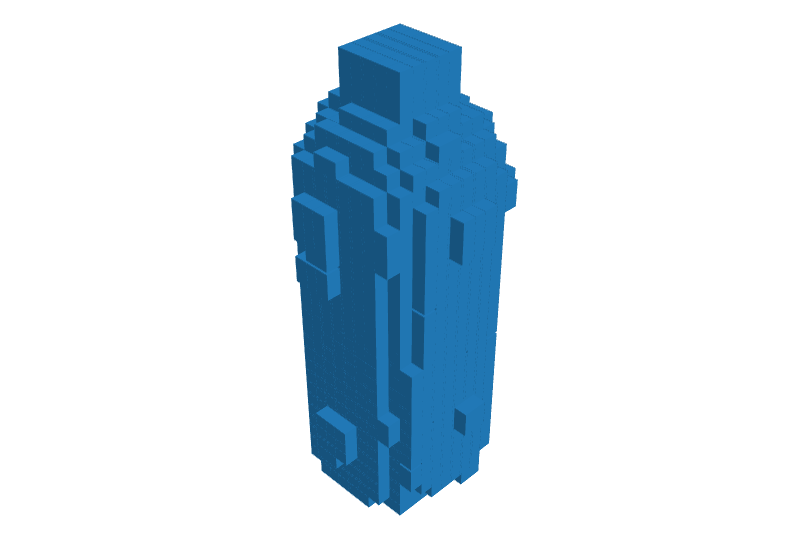}\\
    \includegraphics[width=\dd\textwidth]{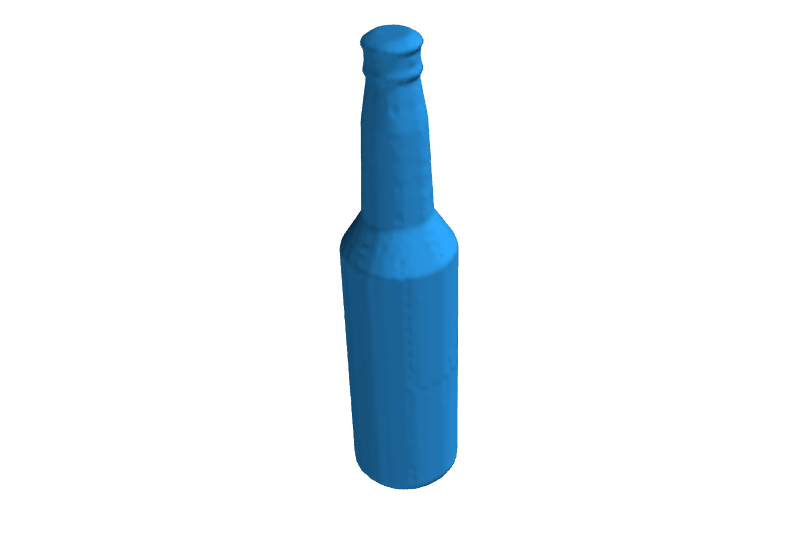}\hspace{1mm} 
    \includegraphics[width=\dd\textwidth]{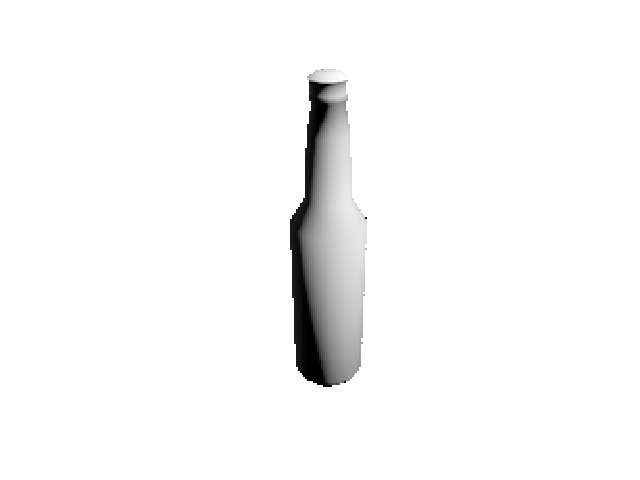}\hspace{1mm} 
    \includegraphics[width=\dd\textwidth]{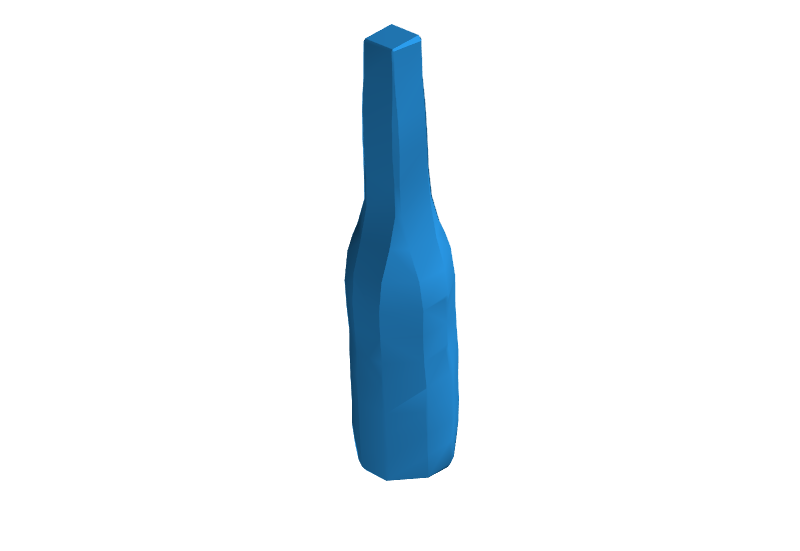}\hspace{1mm}
    \includegraphics[width=\dd\textwidth]{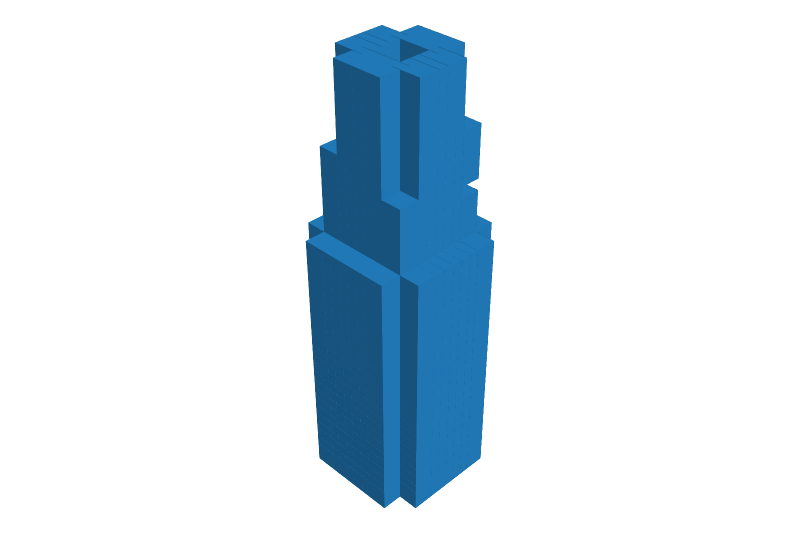}\hspace{1mm}
    \includegraphics[width=\dd\textwidth]{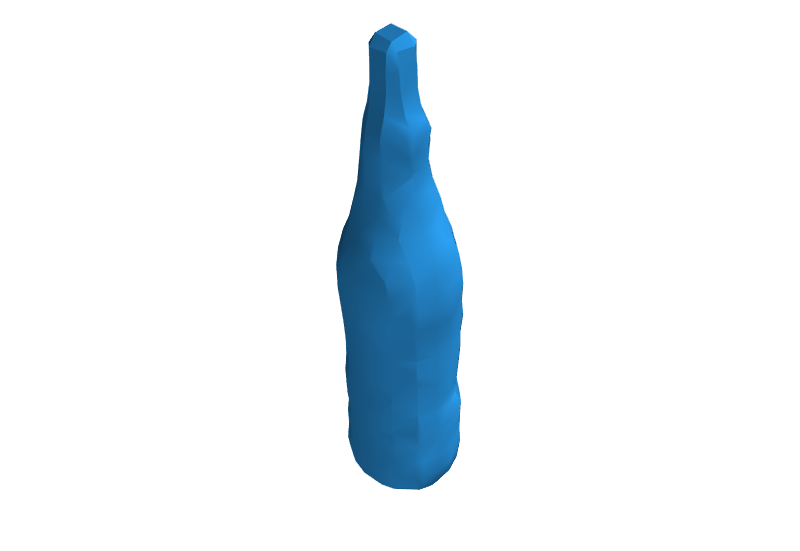}\hspace{1mm}
    \includegraphics[width=\dd\textwidth]{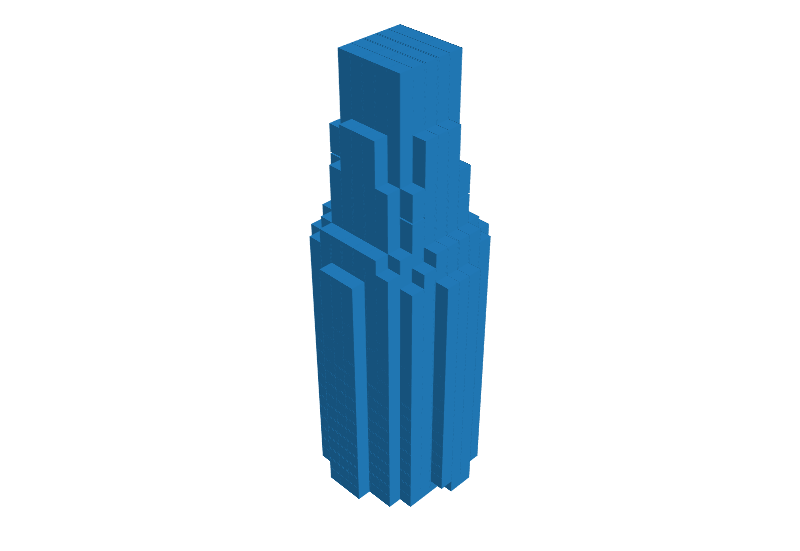}\\
    \includegraphics[width=\dd\textwidth]{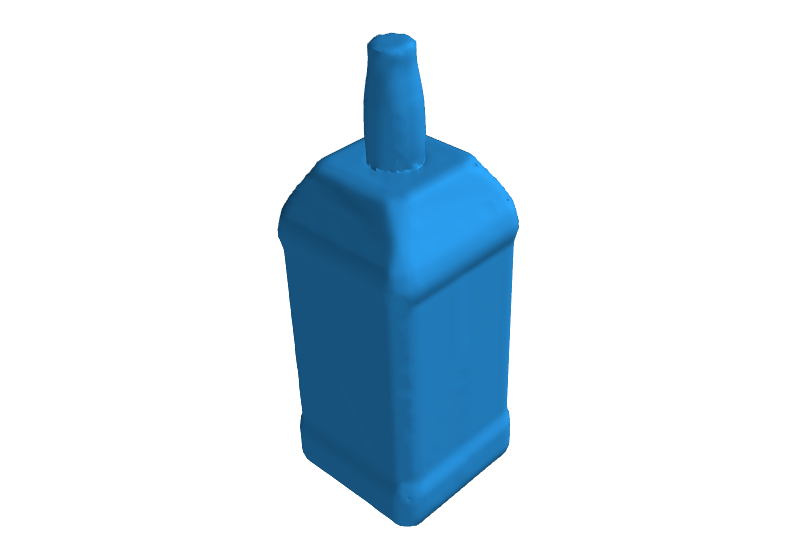}\hspace{1mm} 
    \includegraphics[width=\dd\textwidth]{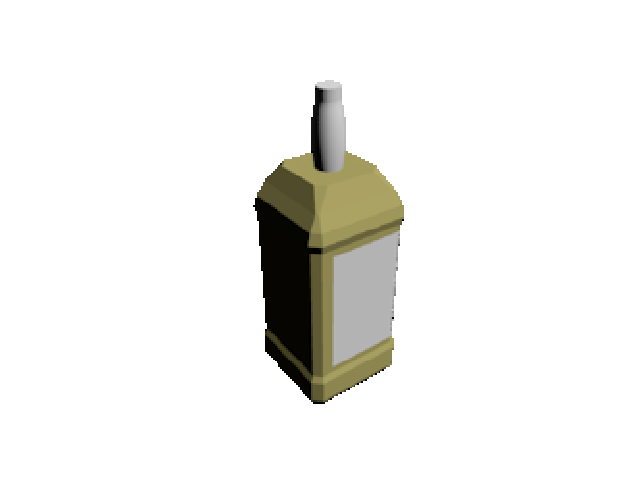}\hspace{1mm} 
    \includegraphics[width=\dd\textwidth]{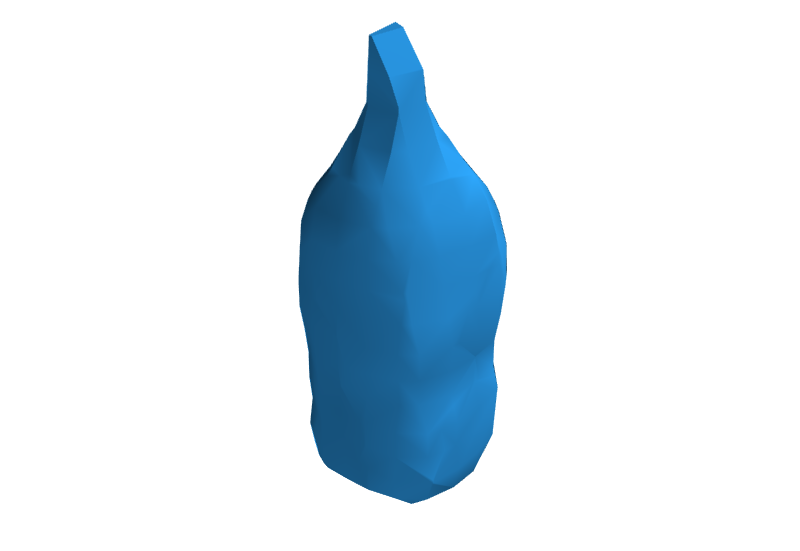}\hspace{1mm}
    \includegraphics[width=\dd\textwidth]{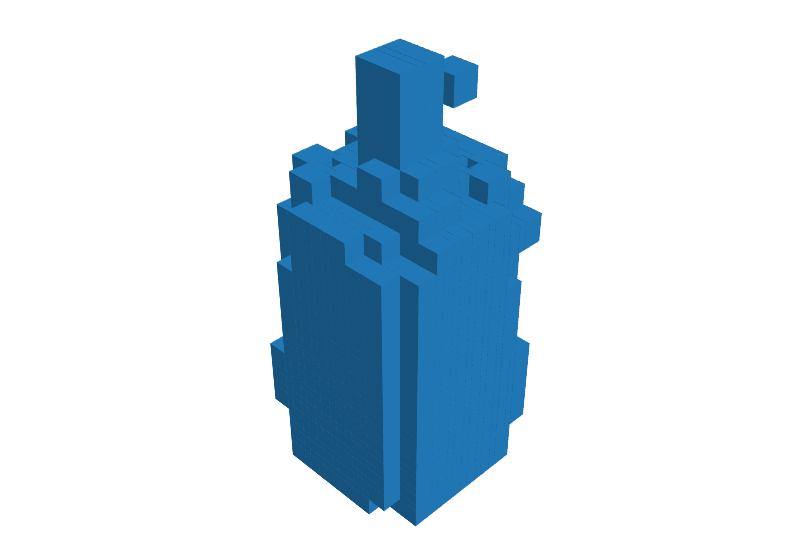}\hspace{1mm}
    \includegraphics[width=\dd\textwidth]{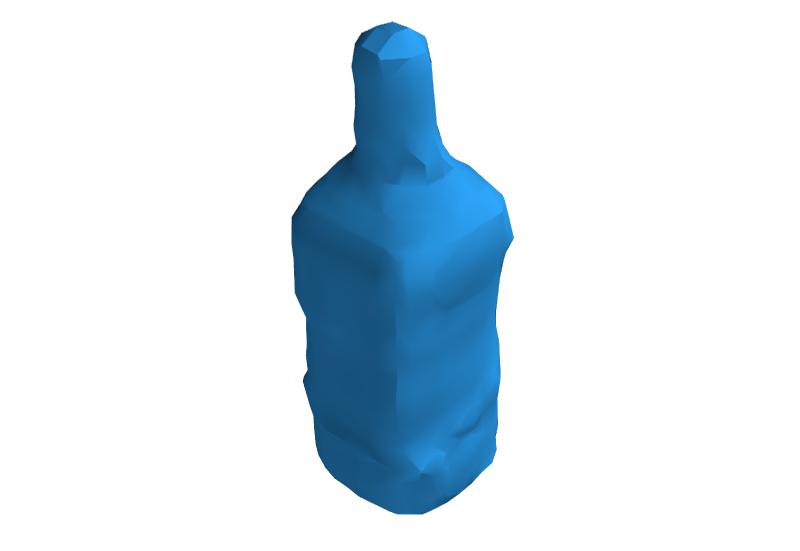}\hspace{1mm}
    \includegraphics[width=\dd\textwidth]{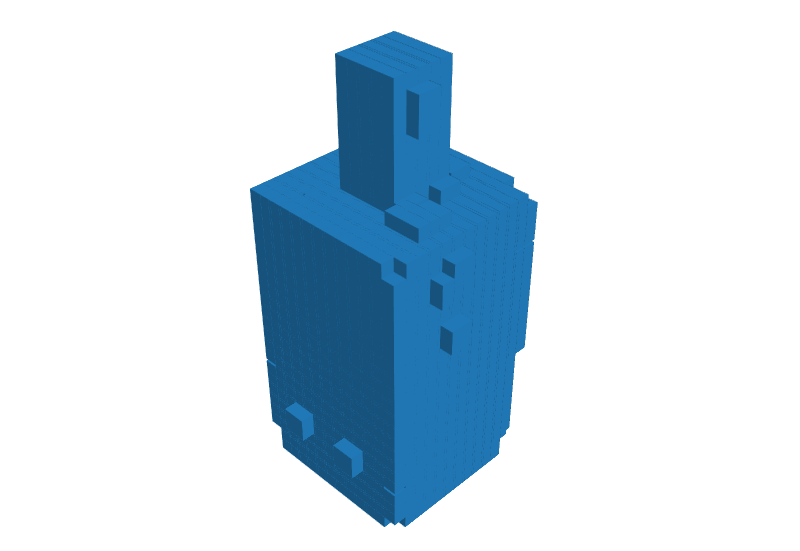}\\
    \includegraphics[width=\dd\textwidth]{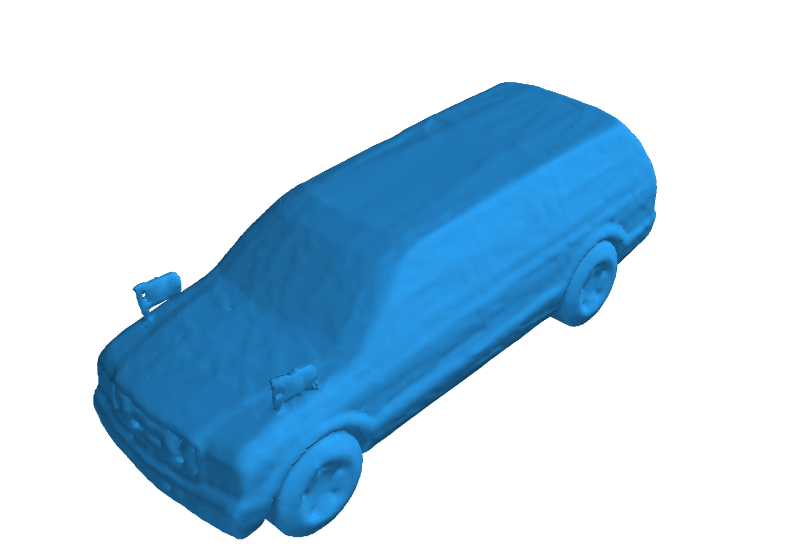}\hspace{1mm} 
    \includegraphics[width=\dd\textwidth]{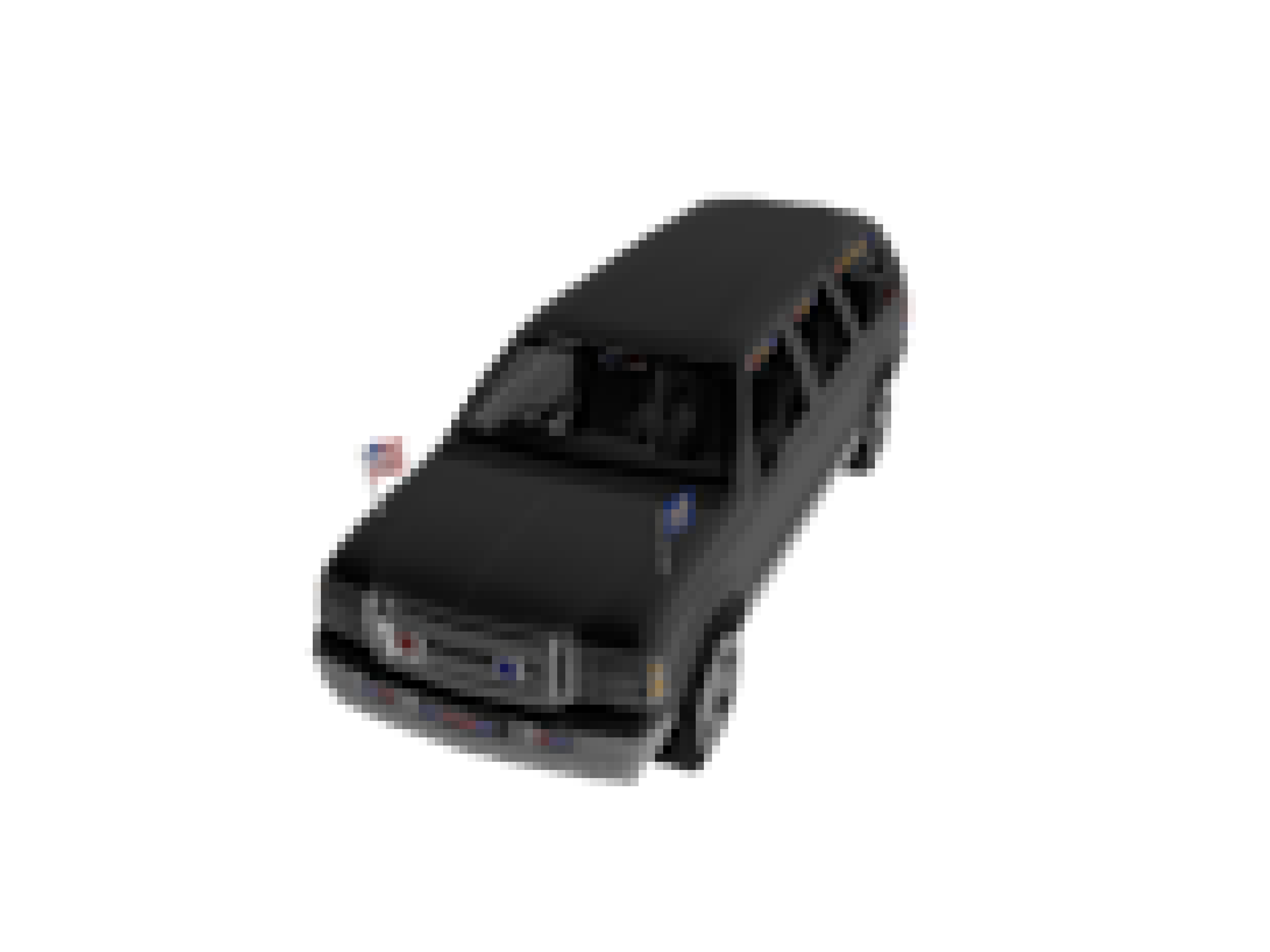}\hspace{1mm} 
    \includegraphics[width=\dd\textwidth]{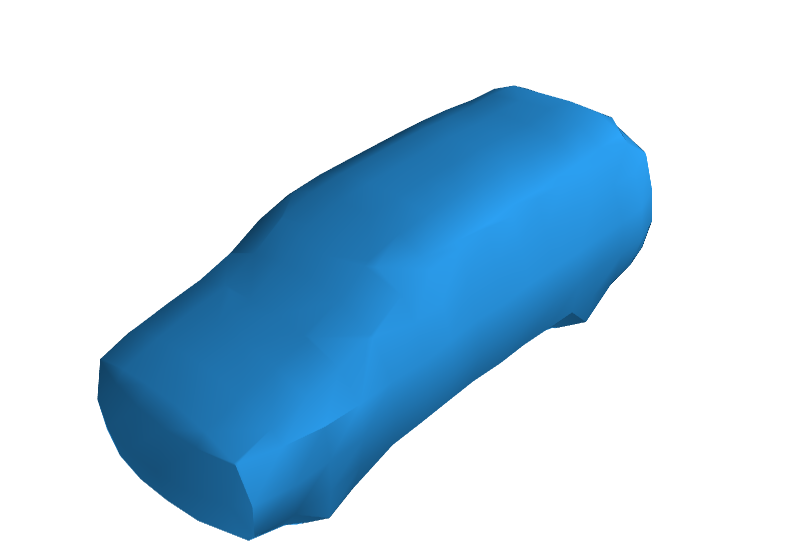}\hspace{1mm}
    \includegraphics[width=\dd\textwidth]{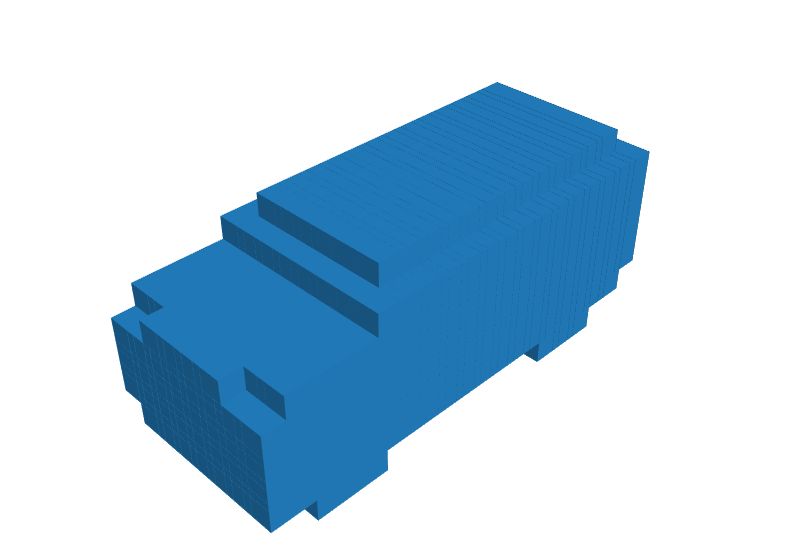}\hspace{1mm}
    \includegraphics[width=\dd\textwidth]{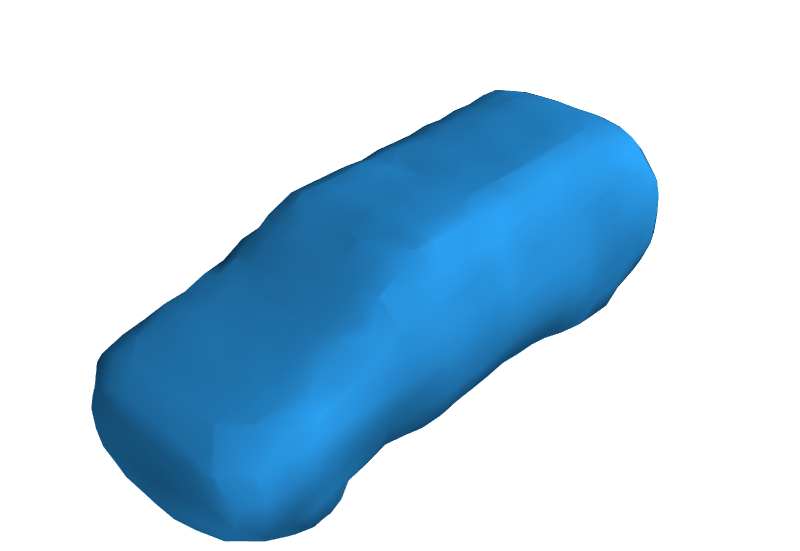}\hspace{1mm}
    \includegraphics[width=\dd\textwidth]{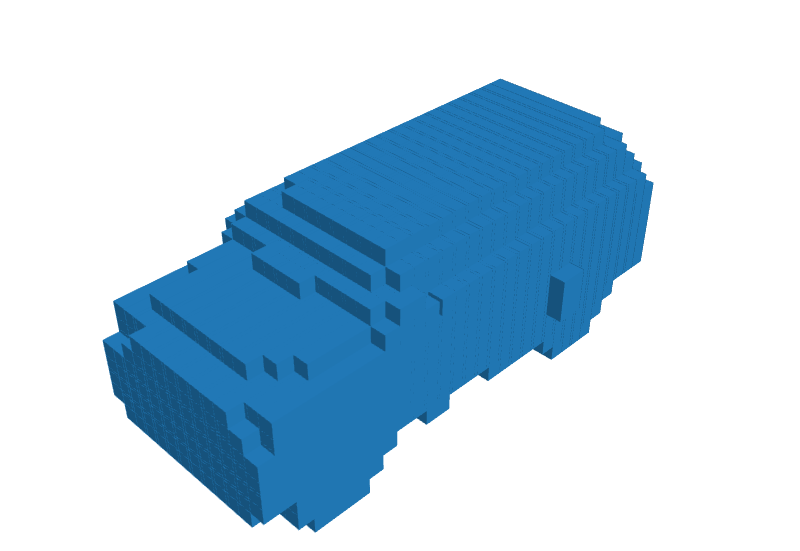}\\
    \includegraphics[width=\dd\textwidth]{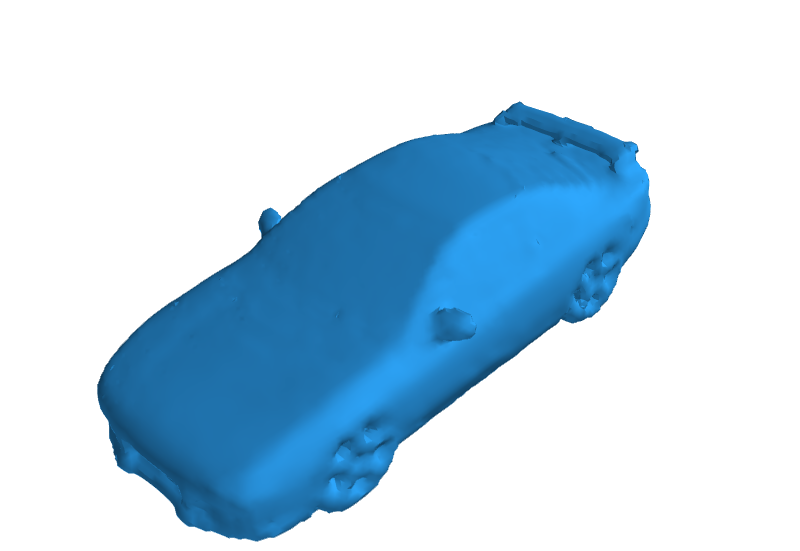}\hspace{1mm} 
    \includegraphics[width=\dd\textwidth]{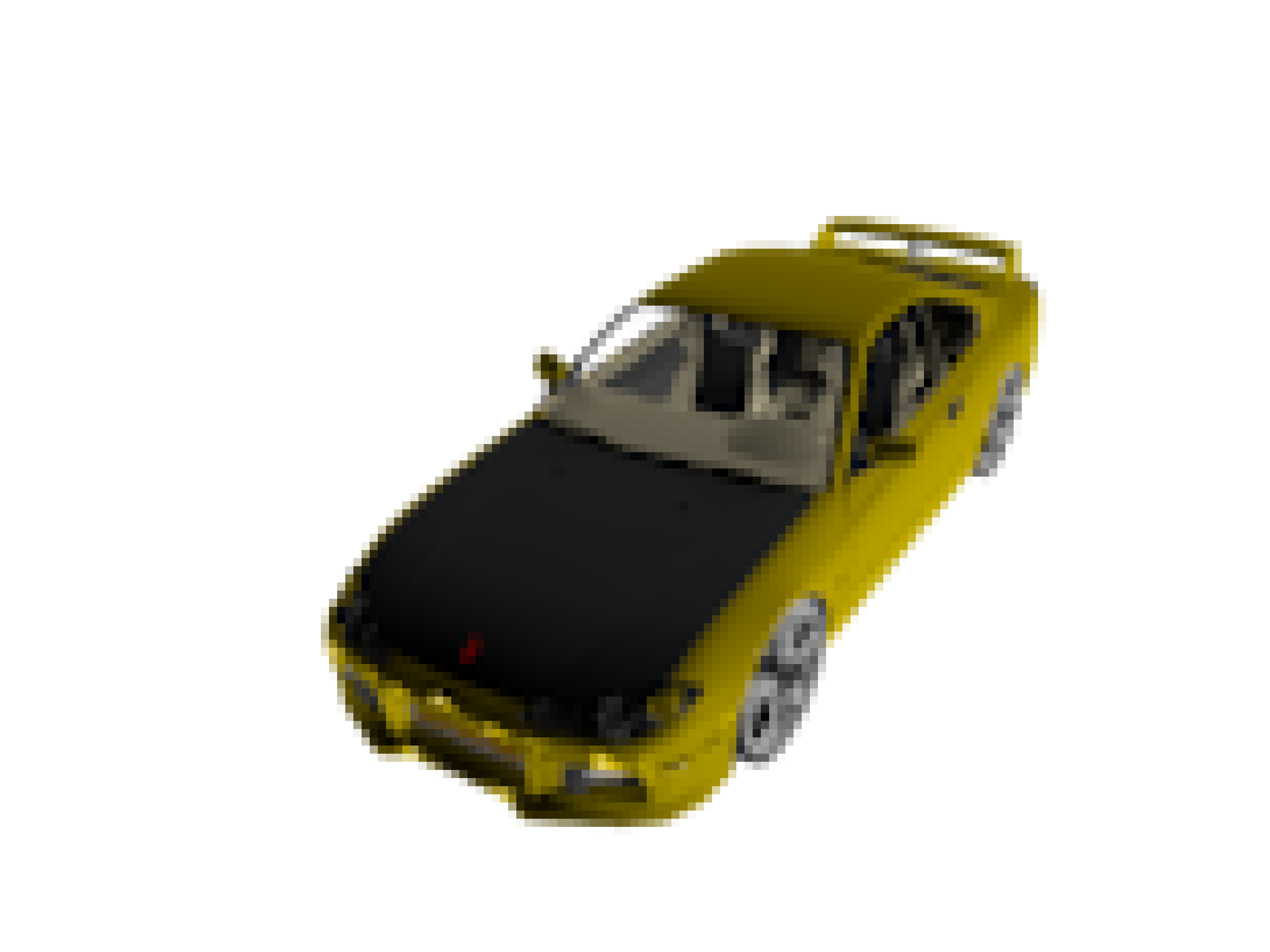}\hspace{1mm} 
    \includegraphics[width=\dd\textwidth]{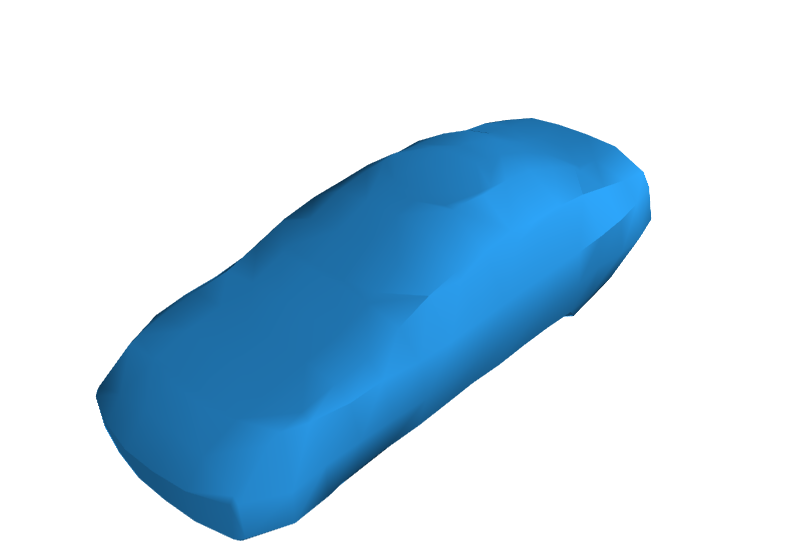}\hspace{1mm}
    \includegraphics[width=\dd\textwidth]{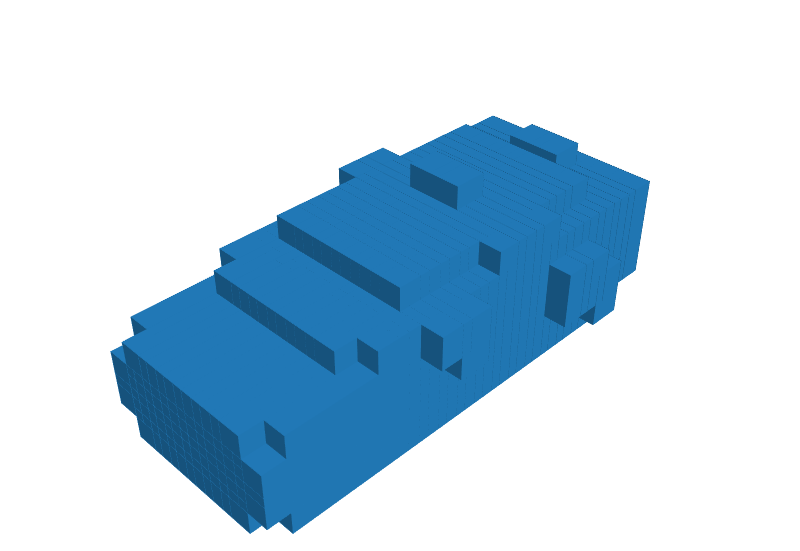}\hspace{1mm}
    \includegraphics[width=\dd\textwidth]{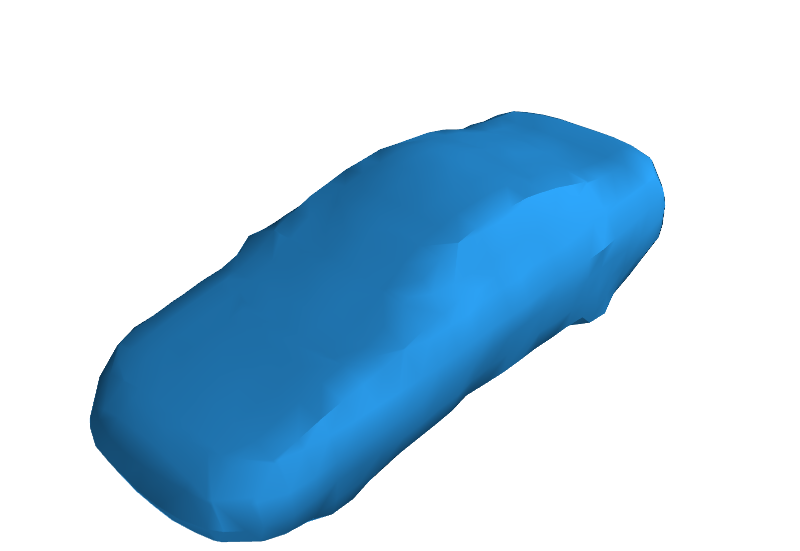}\hspace{1mm}
    \includegraphics[width=\dd\textwidth]{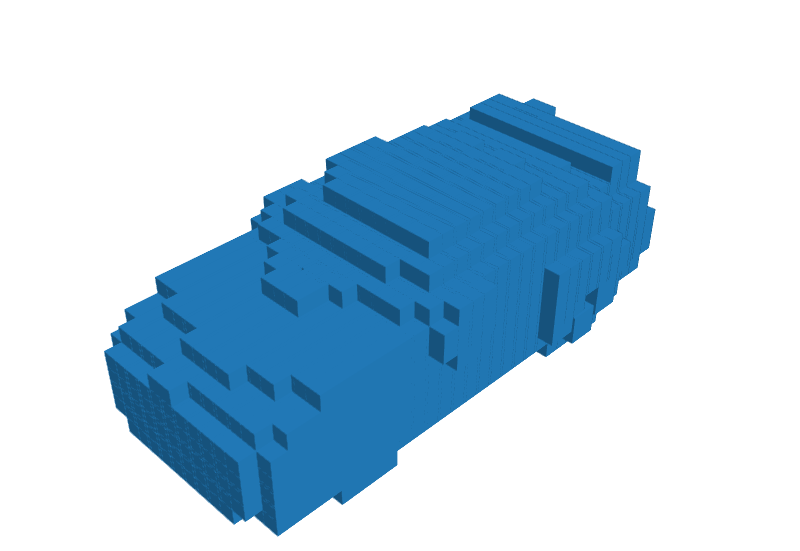}\\
    \includegraphics[width=\dd\textwidth]{./figs/gt6.png}\hspace{1mm} 
    \includegraphics[width=\dd\textwidth]{./figs/im6.png}\hspace{1mm} 
    \includegraphics[width=\dd\textwidth]{./figs/ls20_6.png}\hspace{1mm}
    \includegraphics[width=\dd\textwidth]{./figs/vox20_6.png}\hspace{1mm}
    \includegraphics[width=\dd\textwidth]{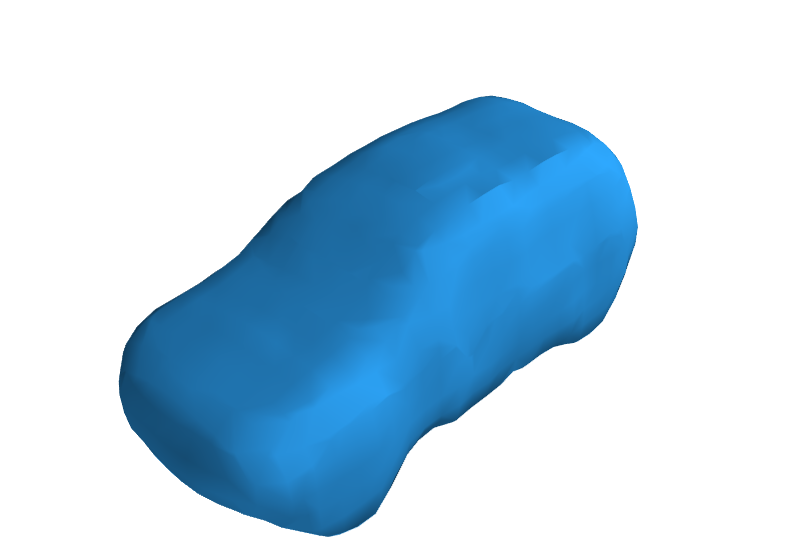}\hspace{1mm}
    \includegraphics[width=\dd\textwidth]{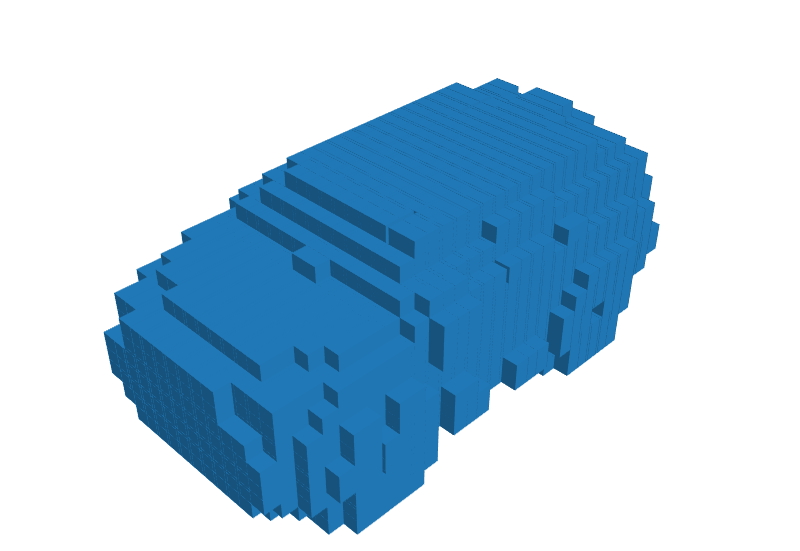}\\
    \includegraphics[width=\dd\textwidth]{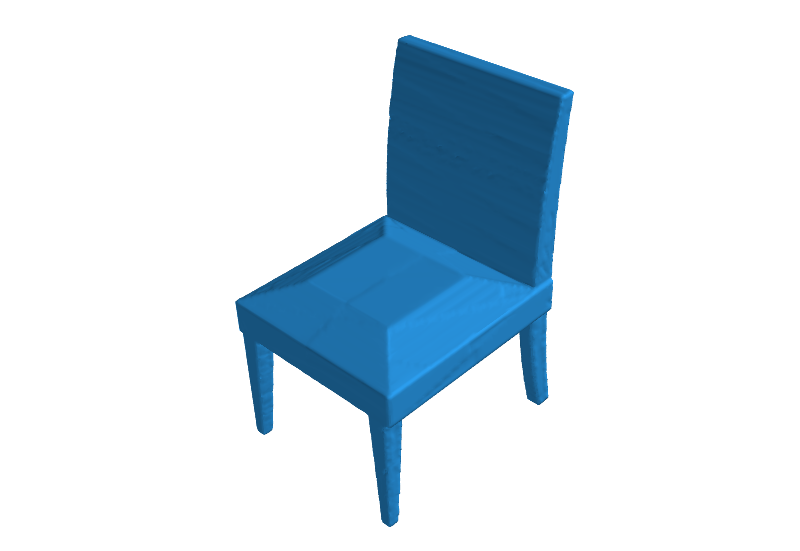}\hspace{1mm} 
    \includegraphics[width=\dd\textwidth]{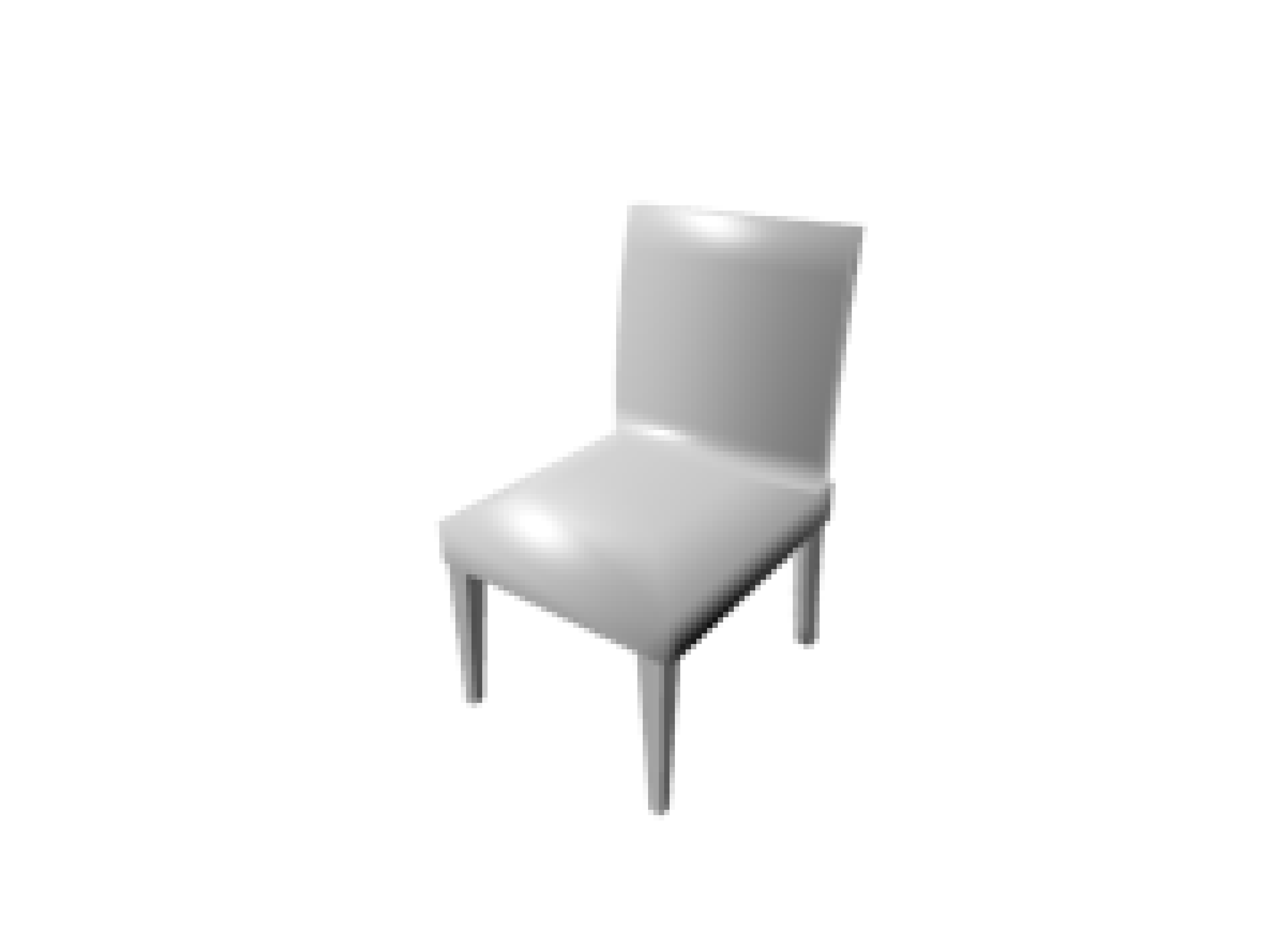}\hspace{1mm} 
    \includegraphics[width=\dd\textwidth]{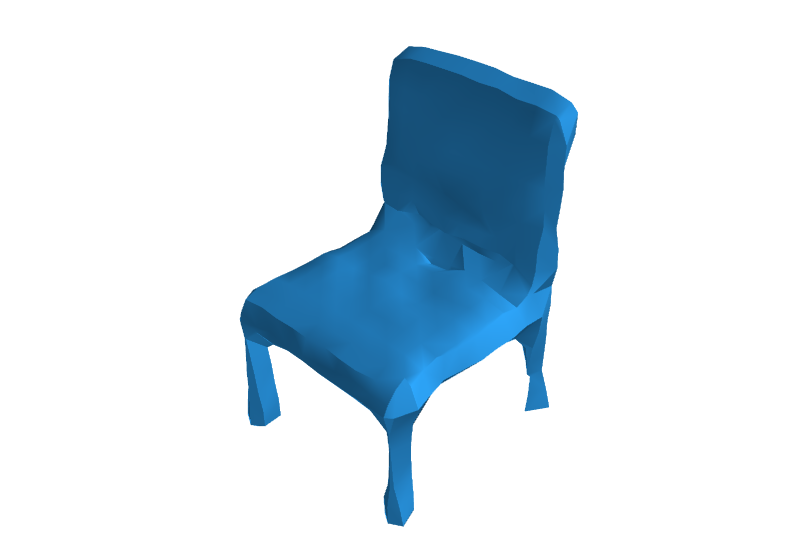}\hspace{1mm}
    \includegraphics[width=\dd\textwidth]{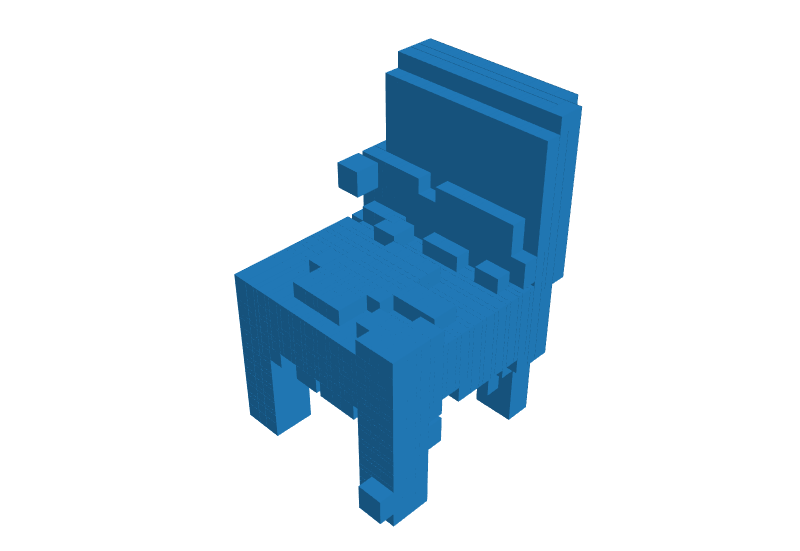}\hspace{1mm}
    \includegraphics[width=\dd\textwidth]{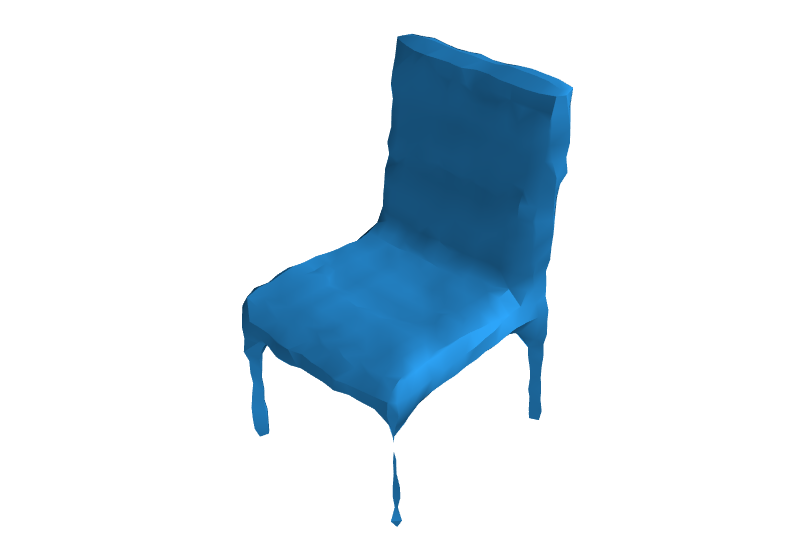}\hspace{1mm}
    \includegraphics[width=\dd\textwidth]{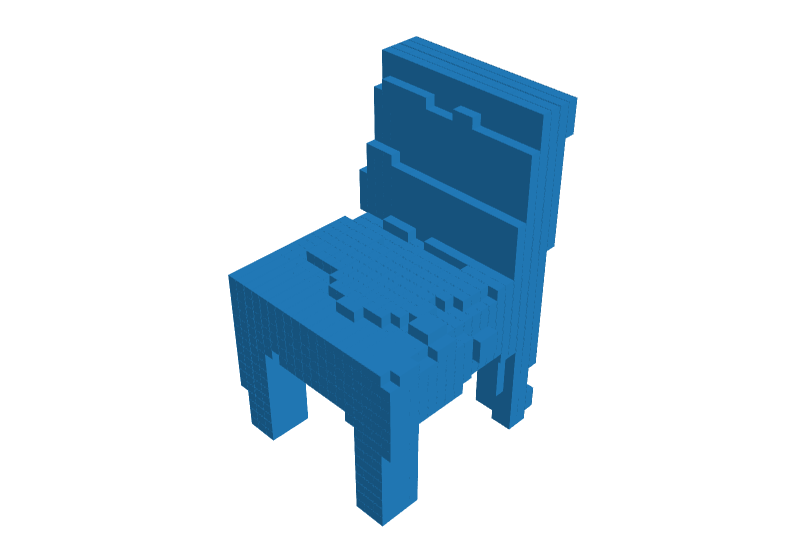}\\
    \includegraphics[width=\dd\textwidth]{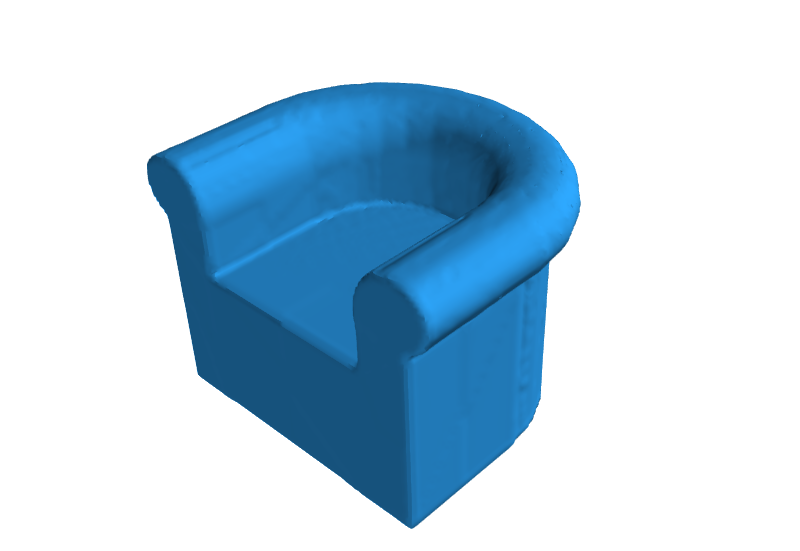}\hspace{1mm} 
    \includegraphics[width=\dd\textwidth]{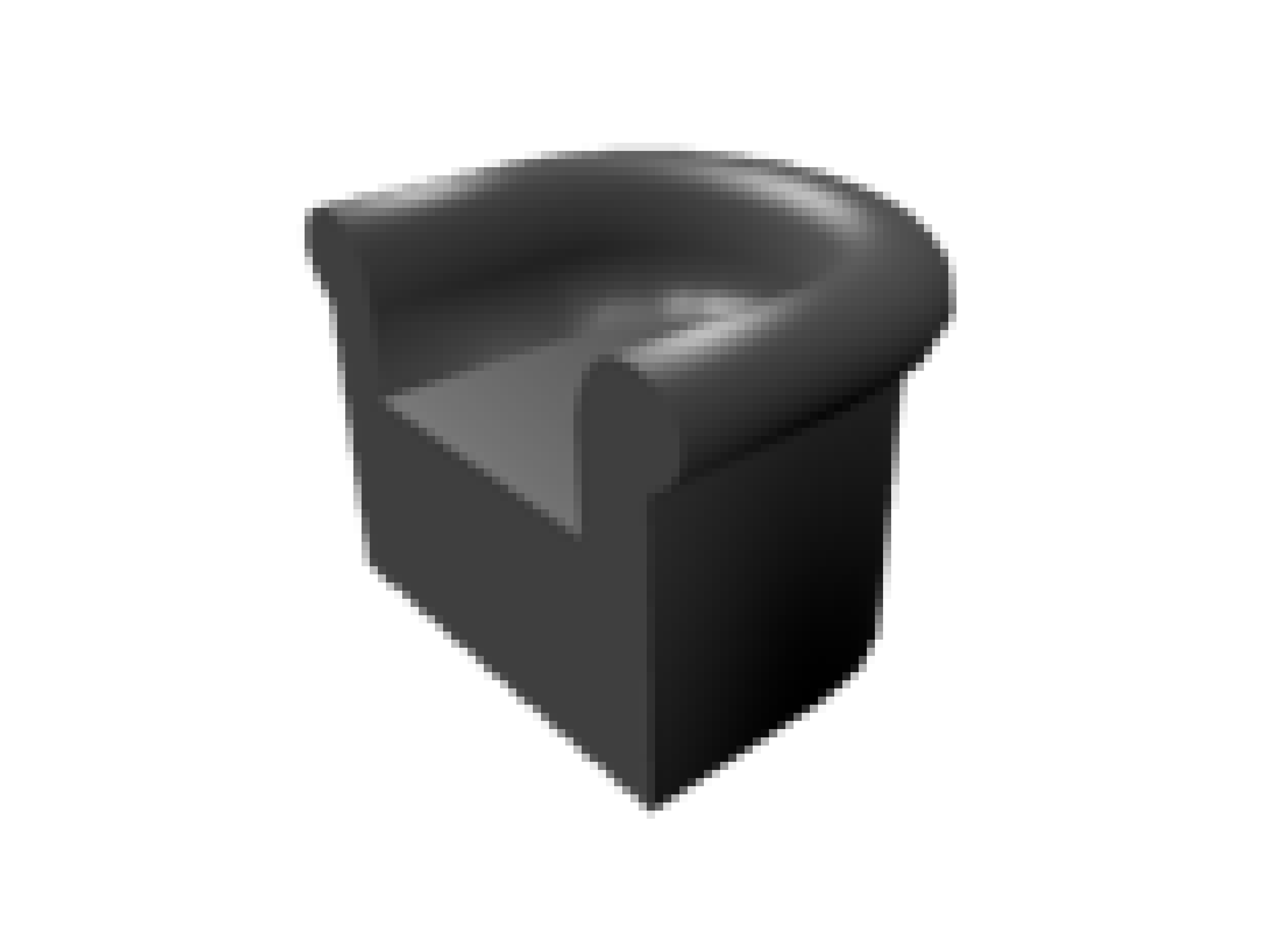}\hspace{1mm} 
    \includegraphics[width=\dd\textwidth]{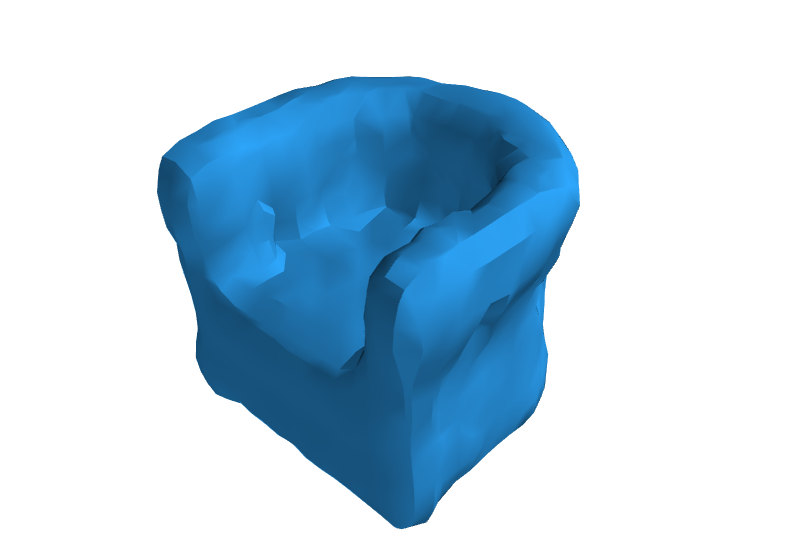}\hspace{1mm}
    \includegraphics[width=\dd\textwidth]{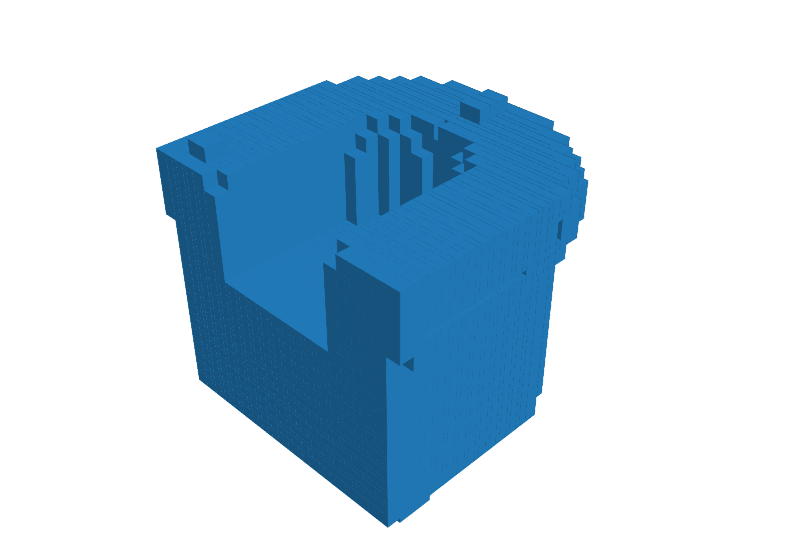}\hspace{1mm}
    \includegraphics[width=\dd\textwidth]{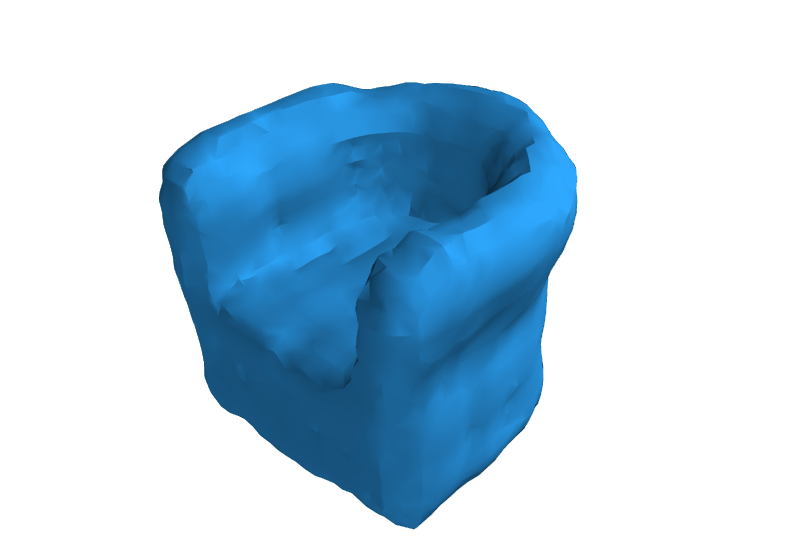}\hspace{1mm}
    \includegraphics[width=\dd\textwidth]{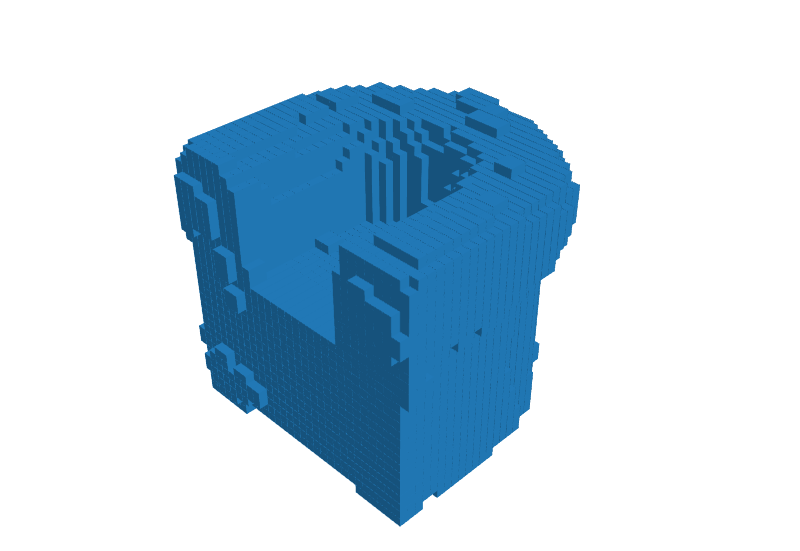}\\
    \includegraphics[width=\dd\textwidth]{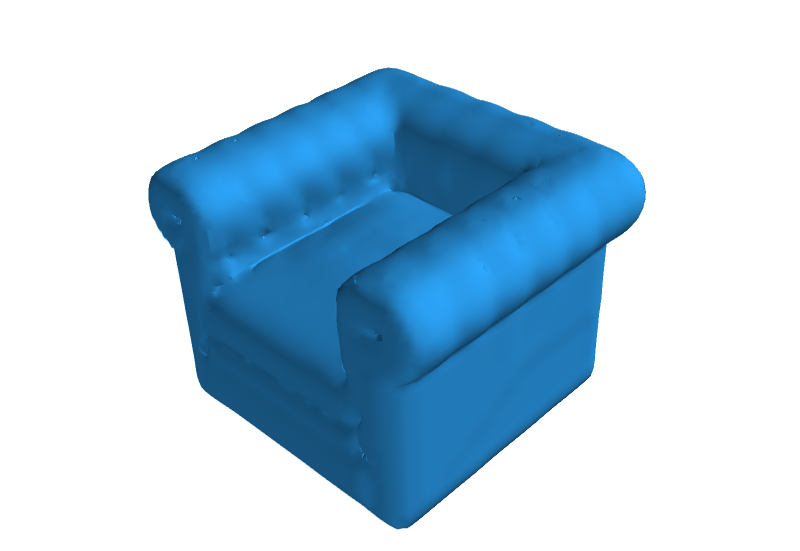}\hspace{1mm} 
    \includegraphics[width=\dd\textwidth]{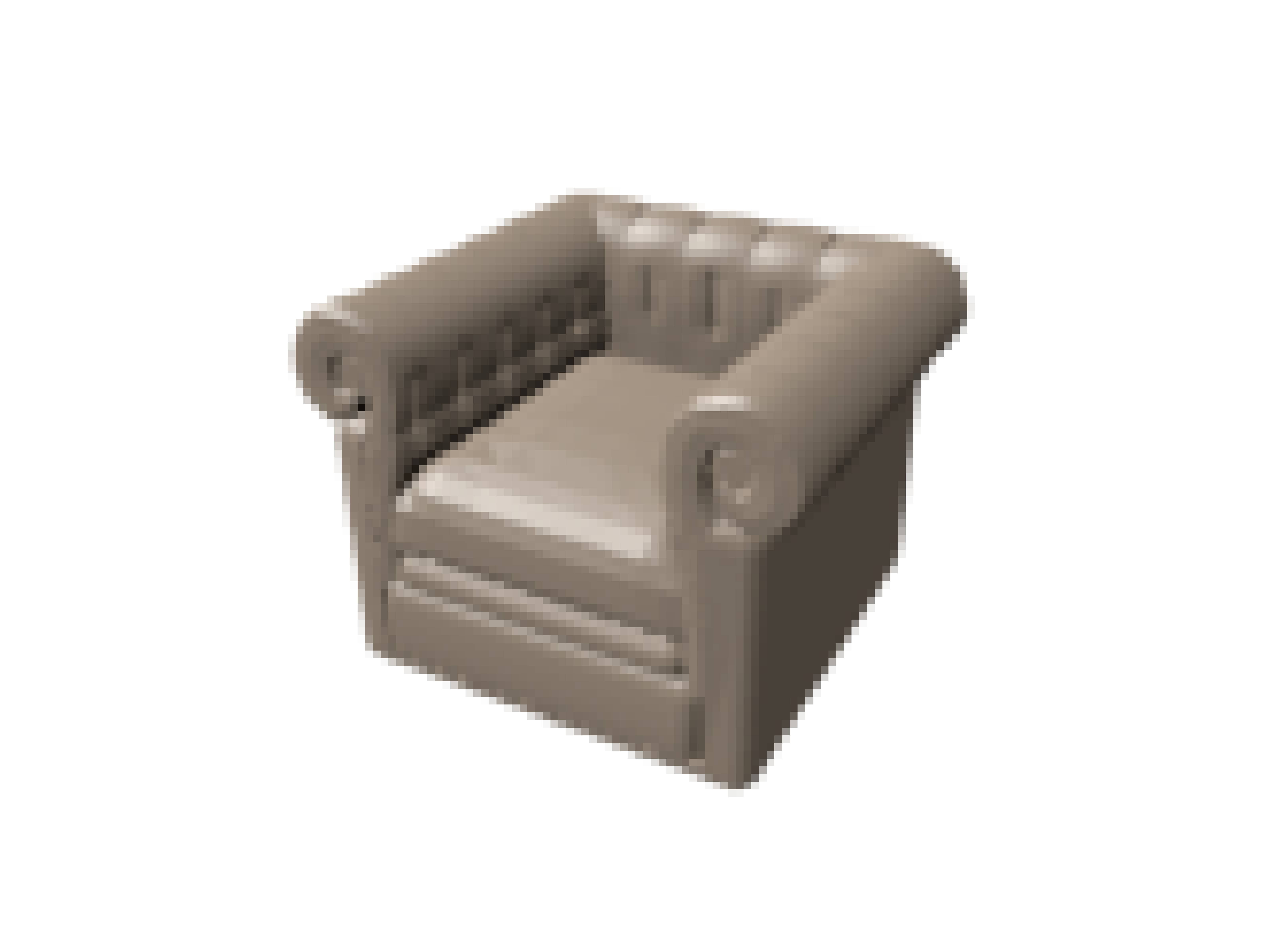}\hspace{1mm} 
    \includegraphics[width=\dd\textwidth]{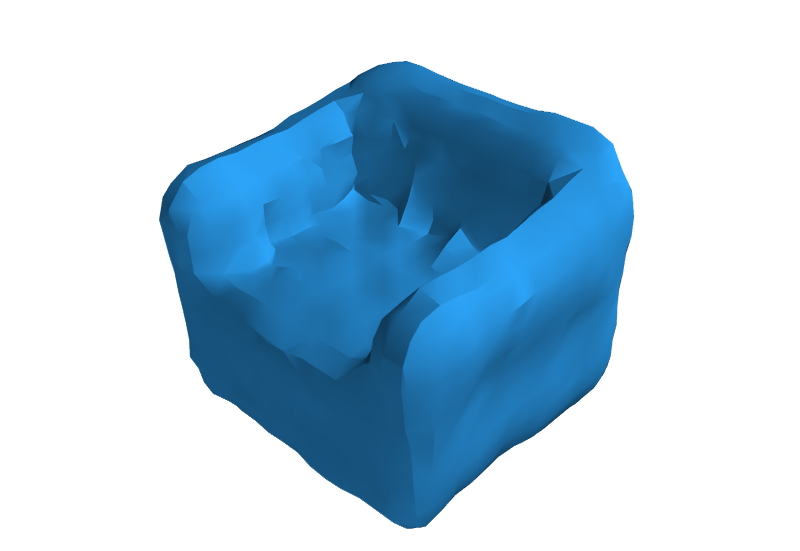}\hspace{1mm}
    \includegraphics[width=\dd\textwidth]{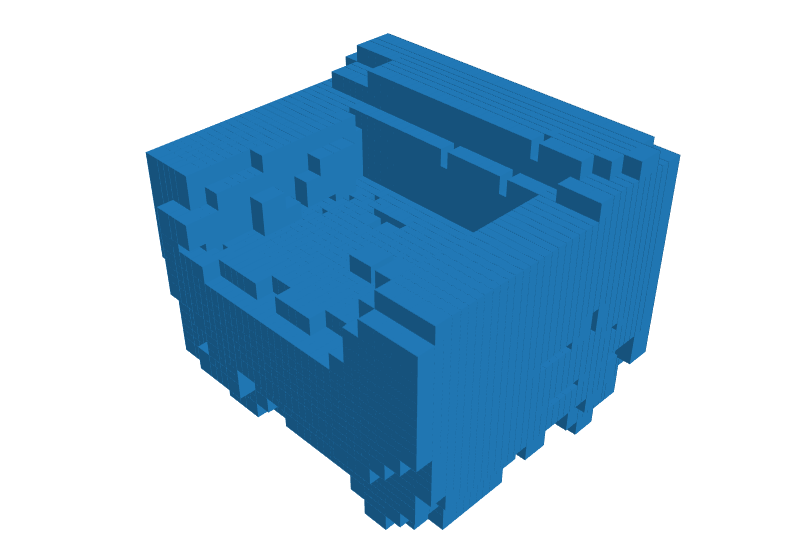}\hspace{1mm}
    \includegraphics[width=\dd\textwidth]{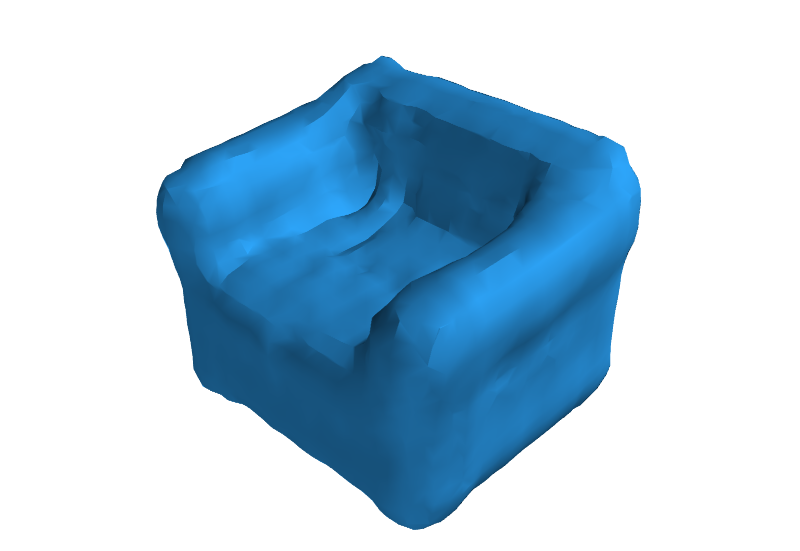}\hspace{1mm}
    \includegraphics[width=\dd\textwidth]{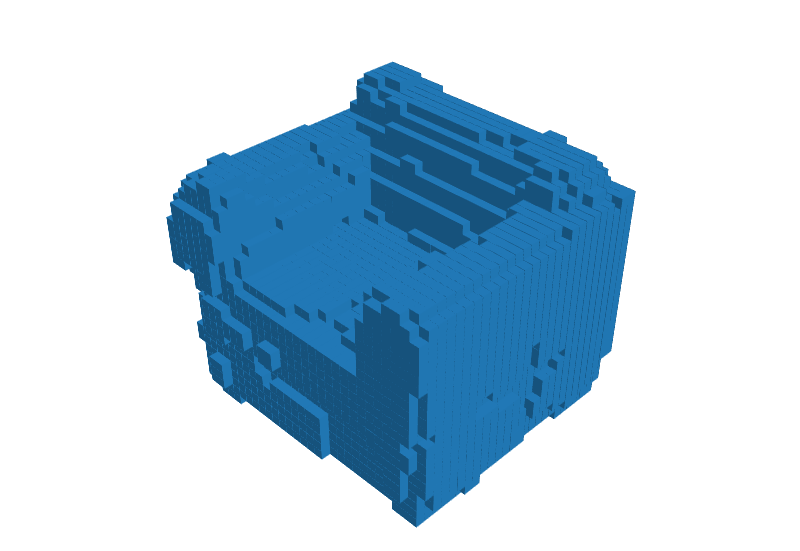}\\
    \includegraphics[width=\dd\textwidth]{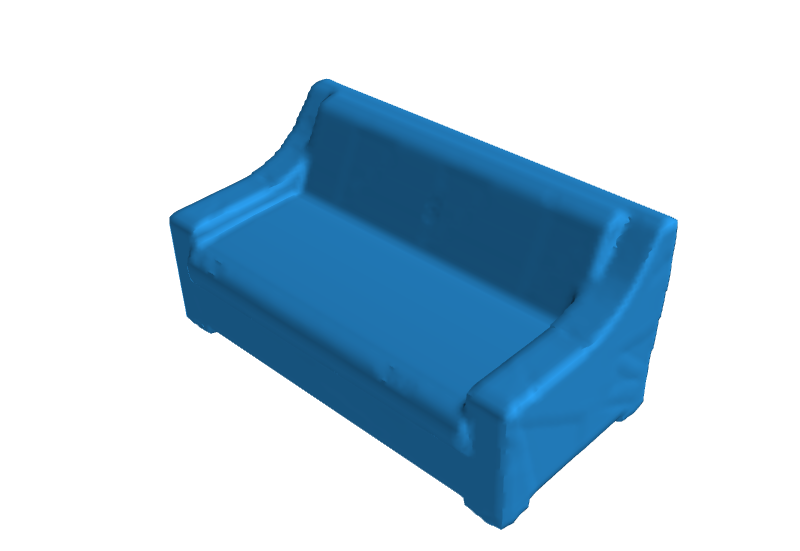}\hspace{1mm} 
    \includegraphics[width=\dd\textwidth]{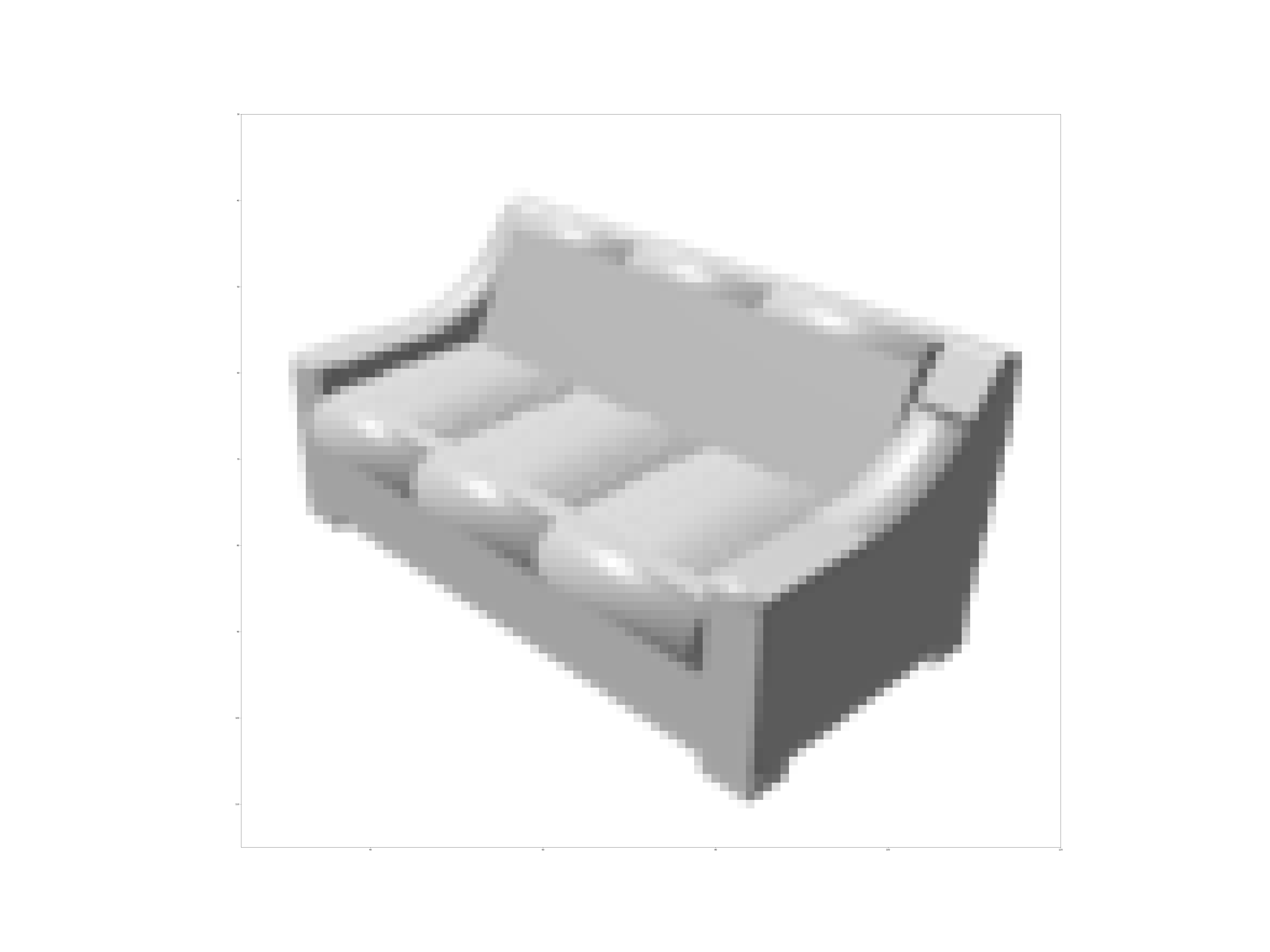}\hspace{1mm} 
    \includegraphics[width=\dd\textwidth]{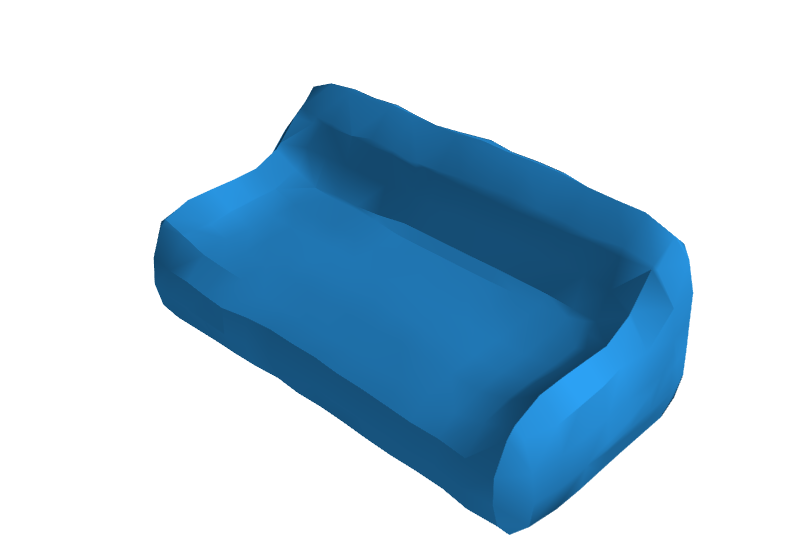}\hspace{1mm}
    \includegraphics[width=\dd\textwidth]{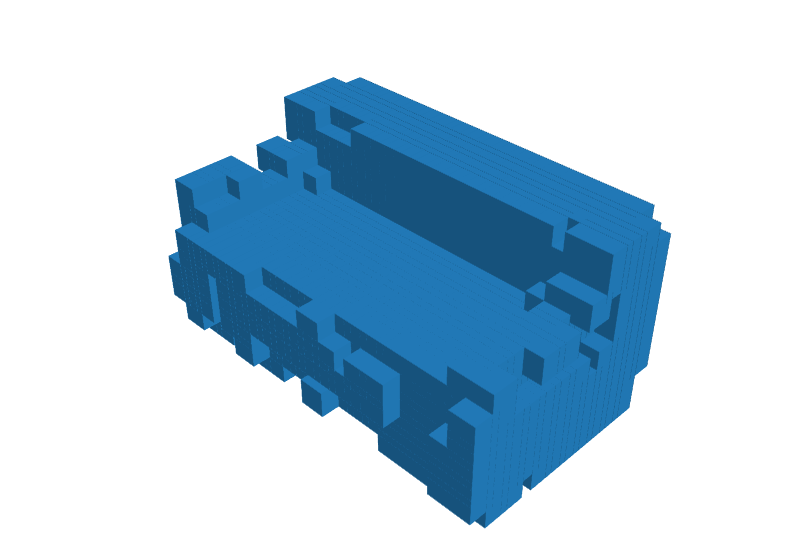}\hspace{1mm}
    \includegraphics[width=\dd\textwidth]{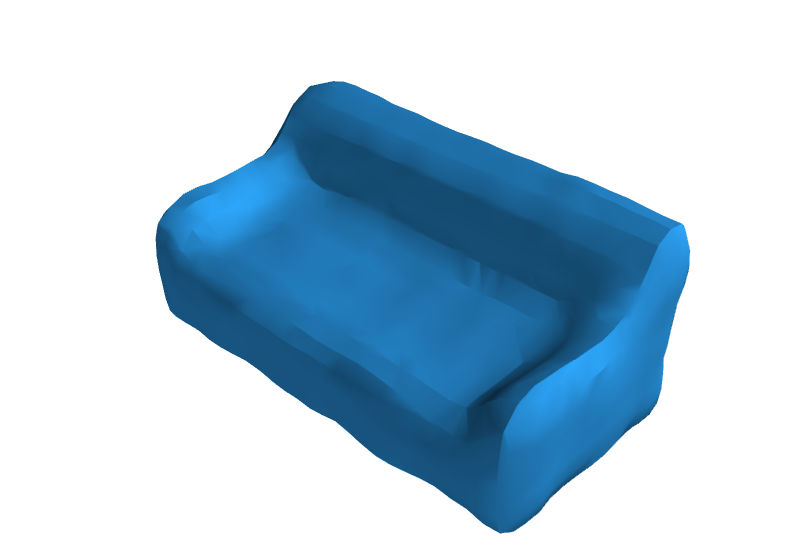}\hspace{1mm}
    \includegraphics[width=\dd\textwidth]{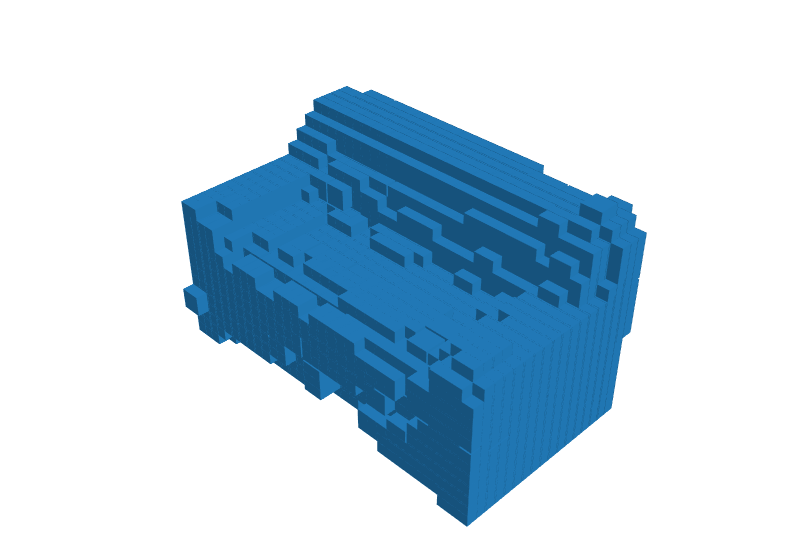}\\
    \includegraphics[width=\dd\textwidth]{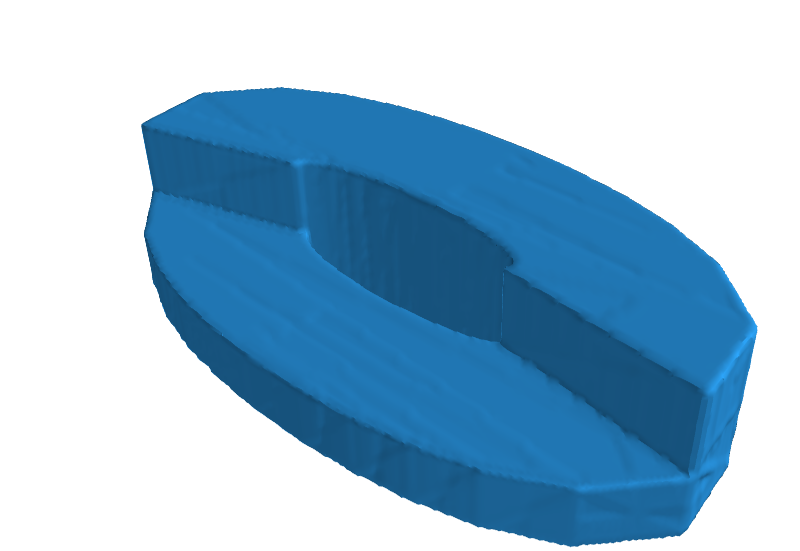}\hspace{1mm} 
    \includegraphics[width=\dd\textwidth]{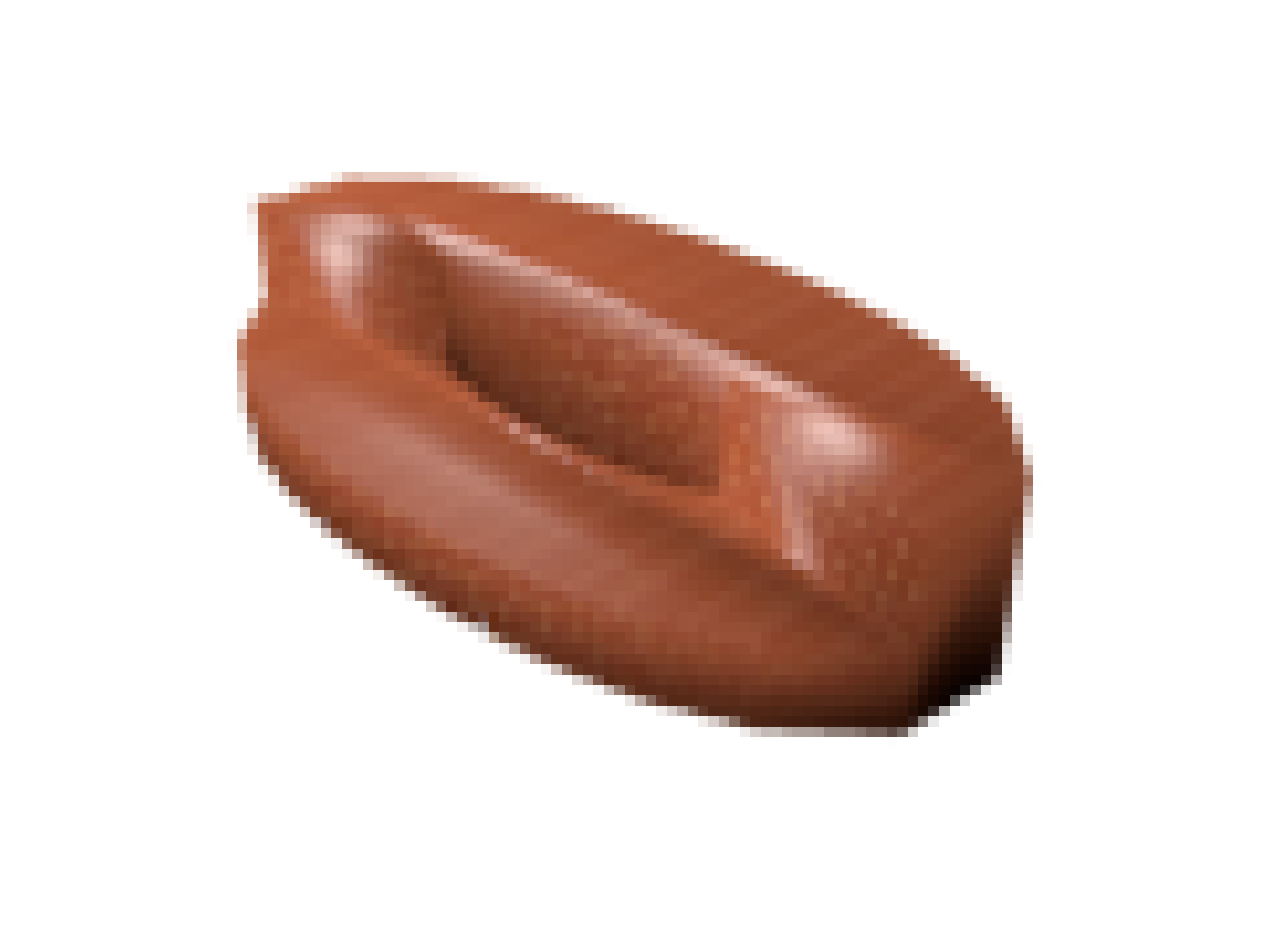}\hspace{1mm} 
    \includegraphics[width=\dd\textwidth]{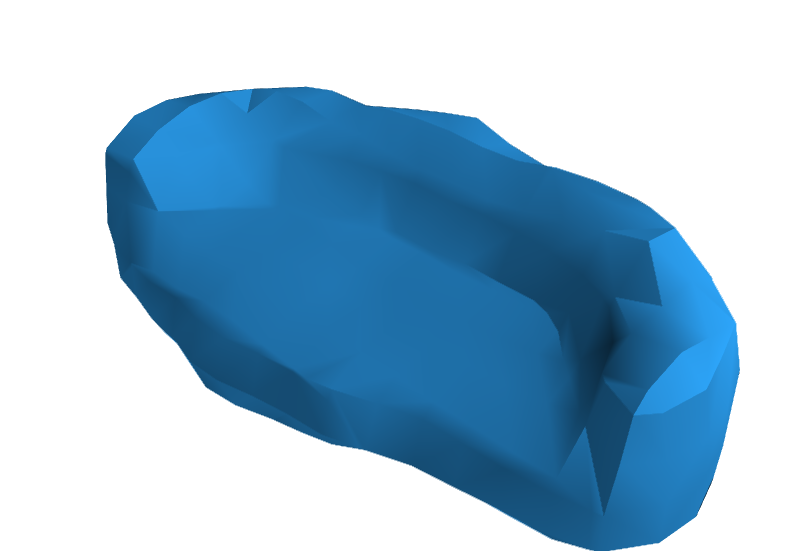}\hspace{1mm}
    \includegraphics[width=\dd\textwidth]{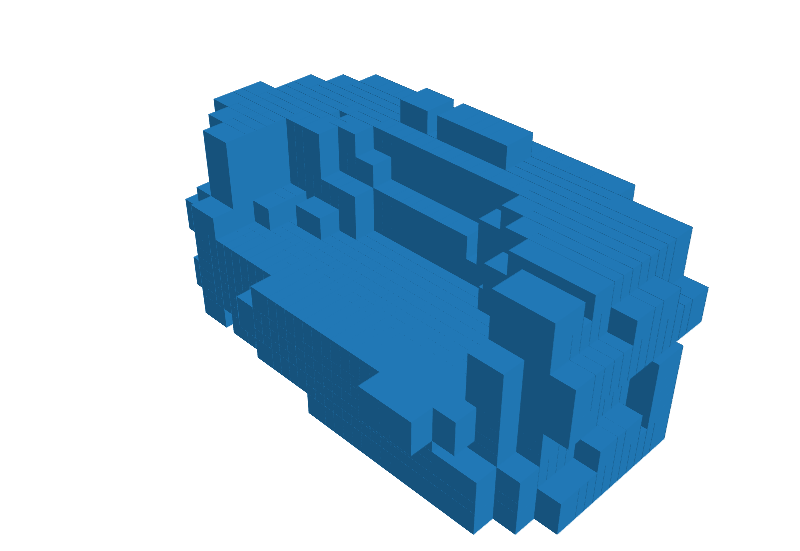}\hspace{1mm}
    \includegraphics[width=\dd\textwidth]{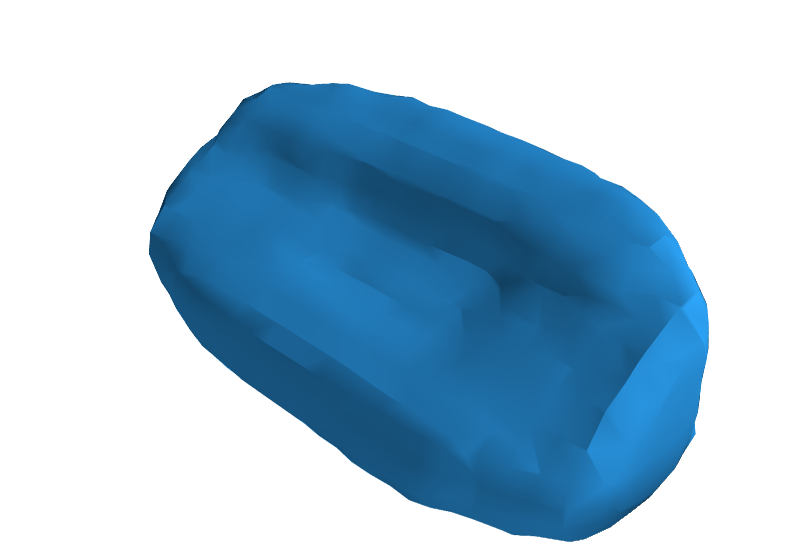}\hspace{1mm}
    \includegraphics[width=\dd\textwidth]{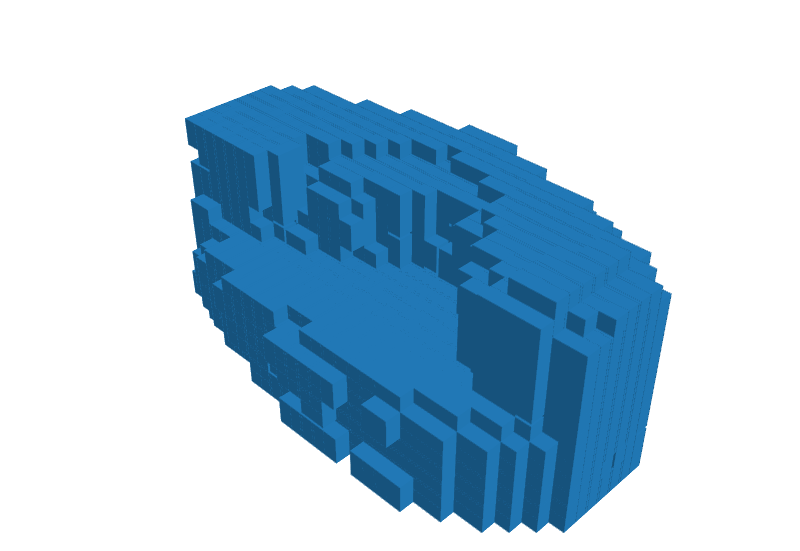}\\
    \includegraphics[width=\dd\textwidth]{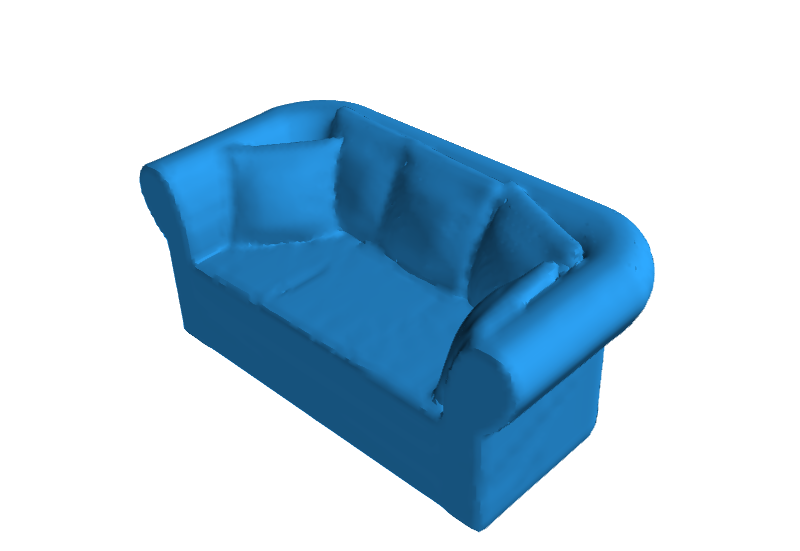}\hspace{1mm} 
    \includegraphics[width=\dd\textwidth]{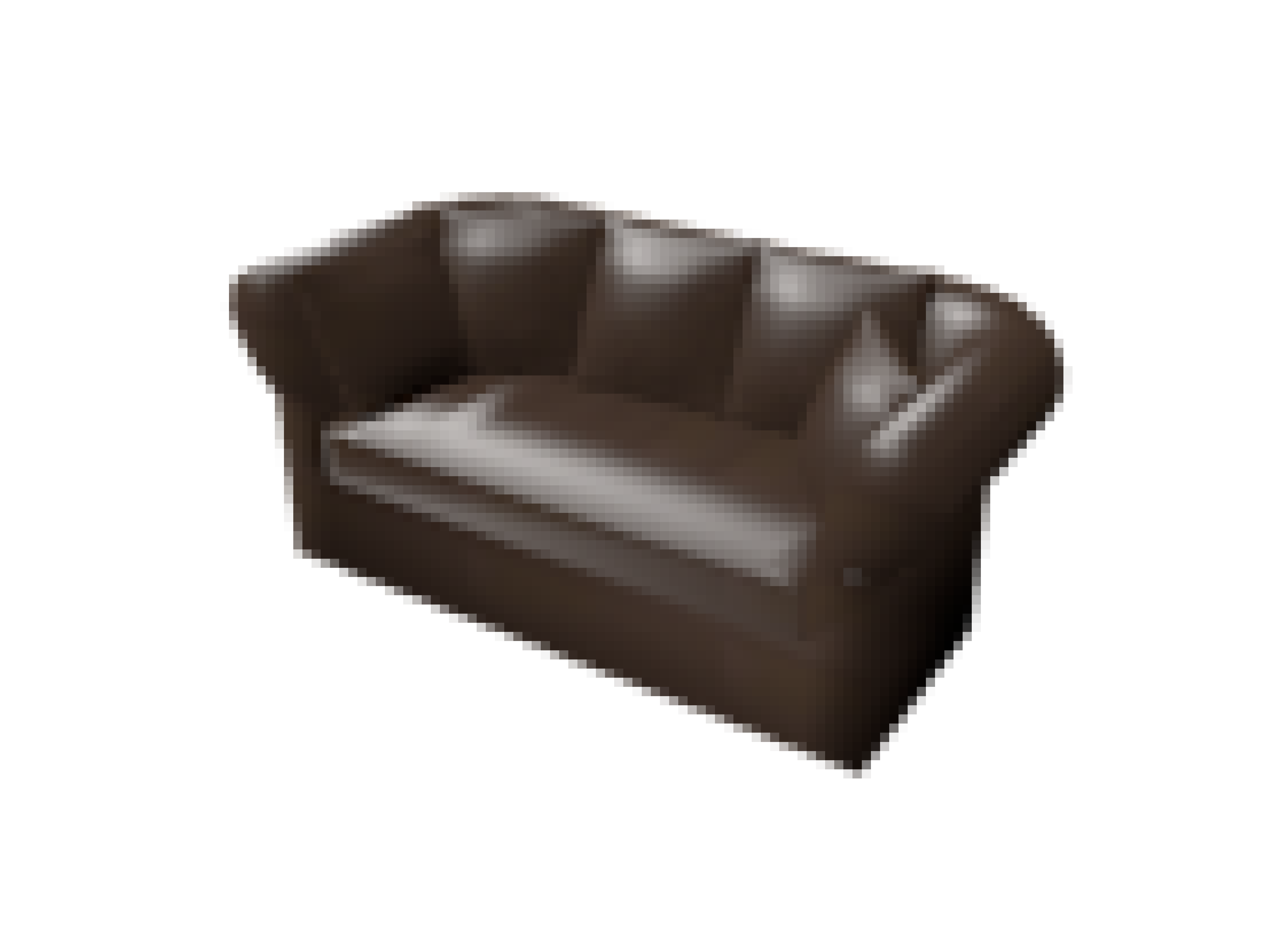}\hspace{1mm} 
    \includegraphics[width=\dd\textwidth]{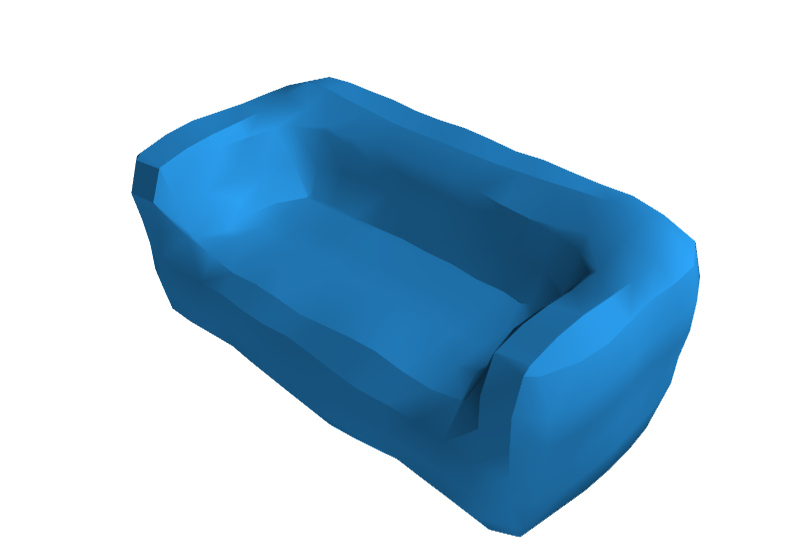}\hspace{1mm}
    \includegraphics[width=\dd\textwidth]{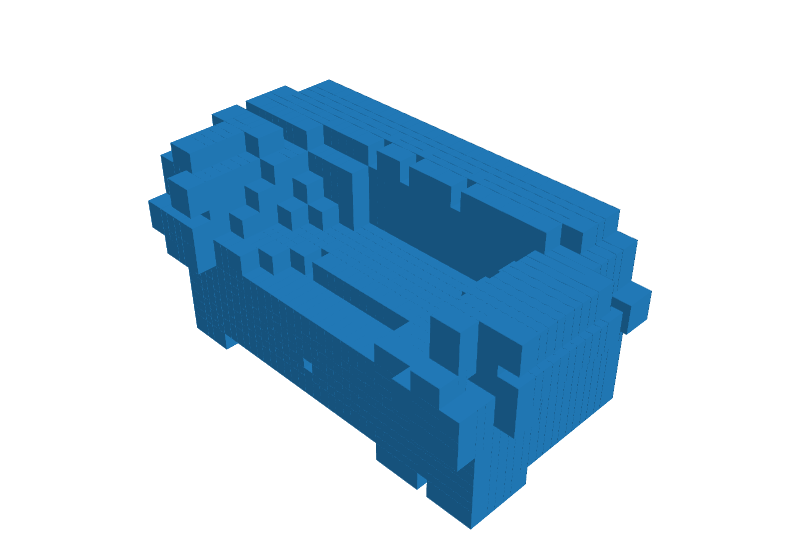}\hspace{1mm}
    \includegraphics[width=\dd\textwidth]{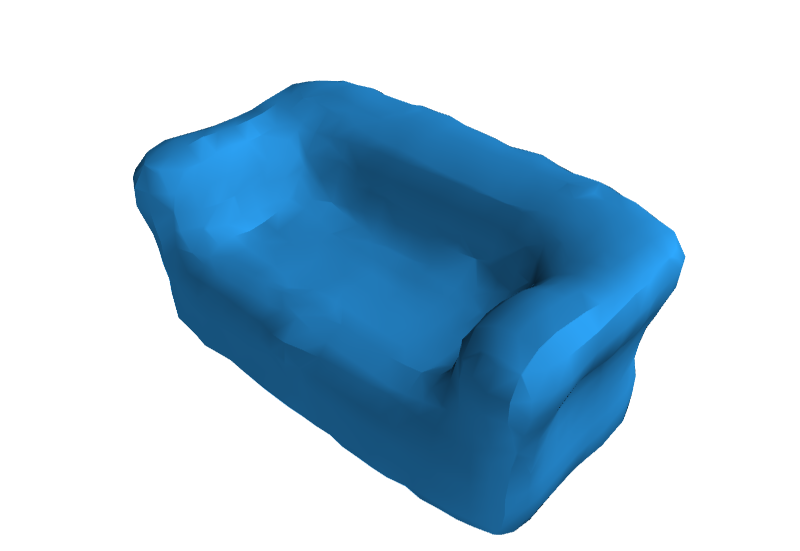}\hspace{1mm}
    \includegraphics[width=\dd\textwidth]{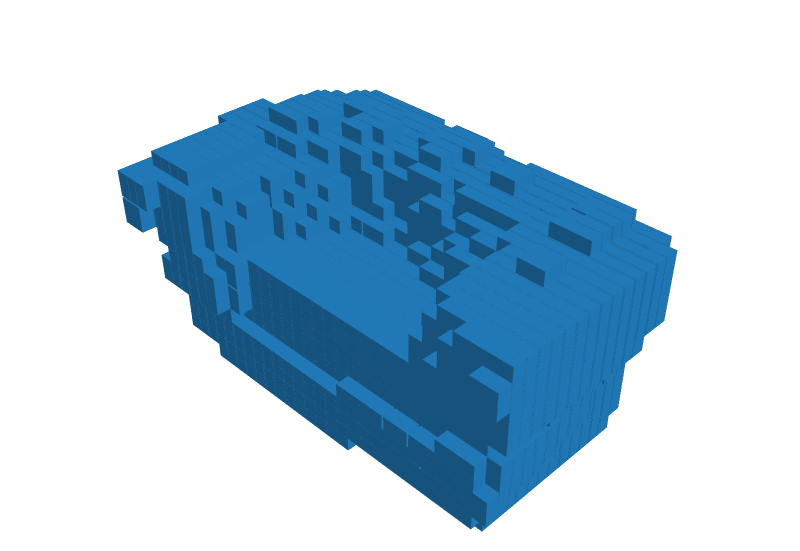}\\
    \caption{ 3D shape inference from a single 2D image. The columns are 
    (a) ground-truth shape, 
    (b) input image, 
    (c) predicted shape, level set, $20^3$, 
    (d)  predicted shape, voxels, $20^3$, 
    (e) predicted shape, level set, $30^3$, 
    (f)  predicted shape, voxels, $30^3$.
    }
    \label{fig:bigone}
\end{figure*}

Examples of the qualitative results are shown in figure \ref{fig:bigone} 
and it clearly demonstrates the higher quality of the 3D shapes inferred by our proposed approach over those obtained from volumetric 
representations.\footnote{These examples 
were chosen to be representative of the typical predictions generated by the different networks.}
%

\section{Conclusion and Future Work}
\label{sec:conclusion}
We proposed a novel and flexible approach for 3D surface prediction by incorporating an oriented variational loss function 
in a deep end-to-end learning framework. We showed that level set functions in its conventional formulation can become ill-conditioned during the evolution process, resulting in numerical and surface motion instabilities. To alleviate this, we made use of an energy functional to promote the unit gradient property as a regulariser in our learning model. Our experiments demonstrated the ability of our approach to infer accurate 3D shapes with more geometrical details compared to voxel-based representations. In future work, we plan to investigate the the flexibility of our approach to accommodate higher resolution shape inference and segmentation problems.

\clearpage

\clearpage

{\small
\bibliographystyle{ieee}
\bibliography{references}
}


\end{document}